\documentclass[pdftex,10pt,journal,final,finalsubmission,twocolumn]{IEEEtran}

\usepackage{graphicx}  

\usepackage{multirow}
\usepackage{pgf}
\usepackage{color}

\definecolor{AliceBlue}{rgb}{0.94,0.97,1.00}
\definecolor{AntiqueWhite1}{rgb}{1.00,0.94,0.86}
\definecolor{AntiqueWhite2}{rgb}{0.93,0.87,0.80}
\definecolor{AntiqueWhite3}{rgb}{0.80,0.75,0.69}
\definecolor{AntiqueWhite4}{rgb}{0.55,0.51,0.47}
\definecolor{AntiqueWhite}{rgb}{0.98,0.92,0.84}
\definecolor{BlanchedAlmond}{rgb}{1.00,0.92,0.80}
\definecolor{BlueViolet}{rgb}{0.54,0.17,0.89}
\definecolor{CadetBlue1}{rgb}{0.60,0.96,1.00}
\definecolor{CadetBlue2}{rgb}{0.56,0.90,0.93}
\definecolor{CadetBlue3}{rgb}{0.48,0.77,0.80}
\definecolor{CadetBlue4}{rgb}{0.33,0.53,0.55}
\definecolor{CadetBlue}{rgb}{0.37,0.62,0.63}
\definecolor{CornflowerBlue}{rgb}{0.39,0.58,0.93}
\definecolor{DarkBlue}{rgb}{0.00,0.00,0.55}
\definecolor{DarkCyan}{rgb}{0.00,0.55,0.55}
\definecolor{DarkGoldenrod1}{rgb}{1.00,0.73,0.06}
\definecolor{DarkGoldenrod2}{rgb}{0.93,0.68,0.05}
\definecolor{DarkGoldenrod3}{rgb}{0.80,0.58,0.05}
\definecolor{DarkGoldenrod4}{rgb}{0.55,0.40,0.03}
\definecolor{DarkGoldenrod}{rgb}{0.72,0.53,0.04}
\definecolor{DarkGray}{rgb}{0.66,0.66,0.66}
\definecolor{DarkGreen}{rgb}{0.00,0.39,0.00}
\definecolor{DarkGrey}{rgb}{0.66,0.66,0.66}
\definecolor{DarkKhaki}{rgb}{0.74,0.72,0.42}
\definecolor{DarkMagenta}{rgb}{0.55,0.00,0.55}
\definecolor{DarkOliveGreen1}{rgb}{0.79,1.00,0.44}
\definecolor{DarkOliveGreen2}{rgb}{0.74,0.93,0.41}
\definecolor{DarkOliveGreen3}{rgb}{0.64,0.80,0.35}
\definecolor{DarkOliveGreen4}{rgb}{0.43,0.55,0.24}
\definecolor{DarkOliveGreen}{rgb}{0.33,0.42,0.18}
\definecolor{OliveGreen}{rgb}{0.33,0.42,0.18}
\definecolor{DarkOrange1}{rgb}{1.00,0.50,0.00}
\definecolor{DarkOrange2}{rgb}{0.93,0.46,0.00}
\definecolor{DarkOrange3}{rgb}{0.80,0.40,0.00}
\definecolor{DarkOrange4}{rgb}{0.55,0.27,0.00}
\definecolor{DarkOrange}{rgb}{1.00,0.55,0.00}
\definecolor{DarkOrchid1}{rgb}{0.75,0.24,1.00}
\definecolor{DarkOrchid2}{rgb}{0.70,0.23,0.93}
\definecolor{DarkOrchid3}{rgb}{0.60,0.20,0.80}
\definecolor{DarkOrchid4}{rgb}{0.41,0.13,0.55}
\definecolor{DarkOrchid}{rgb}{0.60,0.20,0.80}
\definecolor{DarkRed}{rgb}{0.55,0.00,0.00}
\definecolor{DarkSalmon}{rgb}{0.91,0.59,0.48}
\definecolor{DarkSeaGreen1}{rgb}{0.76,1.00,0.76}
\definecolor{DarkSeaGreen2}{rgb}{0.71,0.93,0.71}
\definecolor{DarkSeaGreen3}{rgb}{0.61,0.80,0.61}
\definecolor{DarkSeaGreen4}{rgb}{0.41,0.55,0.41}
\definecolor{DarkSeaGreen}{rgb}{0.56,0.74,0.56}
\definecolor{DarkSlateBlue}{rgb}{0.28,0.24,0.55}
\definecolor{DarkSlateGray1}{rgb}{0.59,1.00,1.00}
\definecolor{DarkSlateGray2}{rgb}{0.55,0.93,0.93}
\definecolor{DarkSlateGray3}{rgb}{0.47,0.80,0.80}
\definecolor{DarkSlateGray4}{rgb}{0.32,0.55,0.55}
\definecolor{DarkSlateGray}{rgb}{0.18,0.31,0.31}
\definecolor{DarkSlateGrey}{rgb}{0.18,0.31,0.31}
\definecolor{DarkTurquoise}{rgb}{0.00,0.81,0.82}
\definecolor{DarkViolet}{rgb}{0.58,0.00,0.83}
\definecolor{DeepPink1}{rgb}{1.00,0.08,0.58}
\definecolor{DeepPink2}{rgb}{0.93,0.07,0.54}
\definecolor{DeepPink3}{rgb}{0.80,0.06,0.46}
\definecolor{DeepPink4}{rgb}{0.55,0.04,0.31}
\definecolor{DeepPink}{rgb}{1.00,0.08,0.58}
\definecolor{DeepSkyBlue1}{rgb}{0.00,0.75,1.00}
\definecolor{DeepSkyBlue2}{rgb}{0.00,0.70,0.93}
\definecolor{DeepSkyBlue3}{rgb}{0.00,0.60,0.80}
\definecolor{DeepSkyBlue4}{rgb}{0.00,0.41,0.55}
\definecolor{DeepSkyBlue}{rgb}{0.00,0.75,1.00}
\definecolor{DimGray}{rgb}{0.41,0.41,0.41}
\definecolor{DimGrey}{rgb}{0.41,0.41,0.41}
\definecolor{DodgerBlue1}{rgb}{0.12,0.56,1.00}
\definecolor{DodgerBlue2}{rgb}{0.11,0.53,0.93}
\definecolor{DodgerBlue3}{rgb}{0.09,0.45,0.80}
\definecolor{DodgerBlue4}{rgb}{0.06,0.31,0.55}
\definecolor{DodgerBlue}{rgb}{0.12,0.56,1.00}
\definecolor{FloralWhite}{rgb}{1.00,0.98,0.94}
\definecolor{ForestGreen}{rgb}{0.13,0.55,0.13}
\definecolor{GhostWhite}{rgb}{0.97,0.97,1.00}
\definecolor{GreenYellow}{rgb}{0.68,1.00,0.18}
\definecolor{HotPink1}{rgb}{1.00,0.43,0.71}
\definecolor{HotPink2}{rgb}{0.93,0.42,0.65}
\definecolor{HotPink3}{rgb}{0.80,0.38,0.56}
\definecolor{HotPink4}{rgb}{0.55,0.23,0.38}
\definecolor{HotPink}{rgb}{1.00,0.41,0.71}
\definecolor{IndianRed1}{rgb}{1.00,0.42,0.42}
\definecolor{IndianRed2}{rgb}{0.93,0.39,0.39}
\definecolor{IndianRed3}{rgb}{0.80,0.33,0.33}
\definecolor{IndianRed4}{rgb}{0.55,0.23,0.23}
\definecolor{IndianRed}{rgb}{0.80,0.36,0.36}
\definecolor{LavenderBlush1}{rgb}{1.00,0.94,0.96}
\definecolor{LavenderBlush2}{rgb}{0.93,0.88,0.90}
\definecolor{LavenderBlush3}{rgb}{0.80,0.76,0.77}
\definecolor{LavenderBlush4}{rgb}{0.55,0.51,0.53}
\definecolor{LavenderBlush}{rgb}{1.00,0.94,0.96}
\definecolor{LawnGreen}{rgb}{0.49,0.99,0.00}
\definecolor{LemonChiffon1}{rgb}{1.00,0.98,0.80}
\definecolor{LemonChiffon2}{rgb}{0.93,0.91,0.75}
\definecolor{LemonChiffon3}{rgb}{0.80,0.79,0.65}
\definecolor{LemonChiffon4}{rgb}{0.55,0.54,0.44}
\definecolor{LemonChiffon}{rgb}{1.00,0.98,0.80}
\definecolor{LightBlue1}{rgb}{0.75,0.94,1.00}
\definecolor{LightBlue2}{rgb}{0.70,0.87,0.93}
\definecolor{LightBlue3}{rgb}{0.60,0.75,0.80}
\definecolor{LightBlue4}{rgb}{0.41,0.51,0.55}
\definecolor{LightBlue}{rgb}{0.68,0.85,0.90}
\definecolor{LightCoral}{rgb}{0.94,0.50,0.50}
\definecolor{LightCyan1}{rgb}{0.88,1.00,1.00}
\definecolor{LightCyan2}{rgb}{0.82,0.93,0.93}
\definecolor{LightCyan3}{rgb}{0.71,0.80,0.80}
\definecolor{LightCyan4}{rgb}{0.48,0.55,0.55}
\definecolor{LightCyan}{rgb}{0.88,1.00,1.00}
\definecolor{LightGoldenrod1}{rgb}{1.00,0.93,0.55}
\definecolor{LightGoldenrod2}{rgb}{0.93,0.86,0.51}
\definecolor{LightGoldenrod3}{rgb}{0.80,0.75,0.44}
\definecolor{LightGoldenrod4}{rgb}{0.55,0.51,0.30}
\definecolor{LightGoldenrodYellow}{rgb}{0.98,0.98,0.82}
\definecolor{LightGoldenrod}{rgb}{0.93,0.87,0.51}
\definecolor{LightGray}{rgb}{0.83,0.83,0.83}
\definecolor{LightGreen}{rgb}{0.56,0.93,0.56}
\definecolor{LightGrey}{rgb}{0.83,0.83,0.83}
\definecolor{LightPink1}{rgb}{1.00,0.68,0.73}
\definecolor{LightPink2}{rgb}{0.93,0.64,0.68}
\definecolor{LightPink3}{rgb}{0.80,0.55,0.58}
\definecolor{LightPink4}{rgb}{0.55,0.37,0.40}
\definecolor{LightPink}{rgb}{1.00,0.71,0.76}
\definecolor{LightSalmon1}{rgb}{1.00,0.63,0.48}
\definecolor{LightSalmon2}{rgb}{0.93,0.58,0.45}
\definecolor{LightSalmon3}{rgb}{0.80,0.51,0.38}
\definecolor{LightSalmon4}{rgb}{0.55,0.34,0.26}
\definecolor{LightSalmon}{rgb}{1.00,0.63,0.48}
\definecolor{LightSeaGreen}{rgb}{0.13,0.70,0.67}
\definecolor{LightSkyBlue1}{rgb}{0.69,0.89,1.00}
\definecolor{LightSkyBlue2}{rgb}{0.64,0.83,0.93}
\definecolor{LightSkyBlue3}{rgb}{0.55,0.71,0.80}
\definecolor{LightSkyBlue4}{rgb}{0.38,0.48,0.55}
\definecolor{LightSkyBlue}{rgb}{0.53,0.81,0.98}
\definecolor{LightSlateBlue}{rgb}{0.52,0.44,1.00}
\definecolor{LightSlateGray}{rgb}{0.47,0.53,0.60}
\definecolor{LightSlateGrey}{rgb}{0.47,0.53,0.60}
\definecolor{LightSteelBlue1}{rgb}{0.79,0.88,1.00}
\definecolor{LightSteelBlue2}{rgb}{0.74,0.82,0.93}
\definecolor{LightSteelBlue3}{rgb}{0.64,0.71,0.80}
\definecolor{LightSteelBlue4}{rgb}{0.43,0.48,0.55}
\definecolor{LightSteelBlue}{rgb}{0.69,0.77,0.87}
\definecolor{LightYellow1}{rgb}{1.00,1.00,0.88}
\definecolor{LightYellow2}{rgb}{0.93,0.93,0.82}
\definecolor{LightYellow3}{rgb}{0.80,0.80,0.71}
\definecolor{LightYellow4}{rgb}{0.55,0.55,0.48}
\definecolor{LightYellow}{rgb}{1.00,1.00,0.88}
\definecolor{LimeGreen}{rgb}{0.20,0.80,0.20}
\definecolor{MediumAquamarine}{rgb}{0.40,0.80,0.67}
\definecolor{MediumBlue}{rgb}{0.00,0.00,0.80}
\definecolor{MediumOrchid1}{rgb}{0.88,0.40,1.00}
\definecolor{MediumOrchid2}{rgb}{0.82,0.37,0.93}
\definecolor{MediumOrchid3}{rgb}{0.71,0.32,0.80}
\definecolor{MediumOrchid4}{rgb}{0.48,0.22,0.55}
\definecolor{MediumOrchid}{rgb}{0.73,0.33,0.83}
\definecolor{MediumPurple1}{rgb}{0.67,0.51,1.00}
\definecolor{MediumPurple2}{rgb}{0.62,0.47,0.93}
\definecolor{MediumPurple3}{rgb}{0.54,0.41,0.80}
\definecolor{MediumPurple4}{rgb}{0.36,0.28,0.55}
\definecolor{MediumPurple}{rgb}{0.58,0.44,0.86}
\definecolor{MediumSeaGreen}{rgb}{0.24,0.70,0.44}
\definecolor{MediumSlateBlue}{rgb}{0.48,0.41,0.93}
\definecolor{MediumSpringGreen}{rgb}{0.00,0.98,0.60}
\definecolor{MediumTurquoise}{rgb}{0.28,0.82,0.80}
\definecolor{MediumVioletRed}{rgb}{0.78,0.08,0.52}
\definecolor{MidnightBlue}{rgb}{0.10,0.10,0.44}
\definecolor{MintCream}{rgb}{0.96,1.00,0.98}
\definecolor{MistyRose1}{rgb}{1.00,0.89,0.88}
\definecolor{MistyRose2}{rgb}{0.93,0.84,0.82}
\definecolor{MistyRose3}{rgb}{0.80,0.72,0.71}
\definecolor{MistyRose4}{rgb}{0.55,0.49,0.48}
\definecolor{MistyRose}{rgb}{1.00,0.89,0.88}
\definecolor{NavajoWhite1}{rgb}{1.00,0.87,0.68}
\definecolor{NavajoWhite2}{rgb}{0.93,0.81,0.63}
\definecolor{NavajoWhite3}{rgb}{0.80,0.70,0.55}
\definecolor{NavajoWhite4}{rgb}{0.55,0.47,0.37}
\definecolor{NavajoWhite}{rgb}{1.00,0.87,0.68}
\definecolor{NavyBlue}{rgb}{0.00,0.00,0.50}
\definecolor{OldLace}{rgb}{0.99,0.96,0.90}
\definecolor{OliveDrab1}{rgb}{0.75,1.00,0.24}
\definecolor{OliveDrab2}{rgb}{0.70,0.93,0.23}
\definecolor{OliveDrab3}{rgb}{0.60,0.80,0.20}
\definecolor{OliveDrab4}{rgb}{0.41,0.55,0.13}
\definecolor{OliveDrab}{rgb}{0.42,0.56,0.14}
\definecolor{OrangeRed1}{rgb}{1.00,0.27,0.00}
\definecolor{OrangeRed2}{rgb}{0.93,0.25,0.00}
\definecolor{OrangeRed3}{rgb}{0.80,0.22,0.00}
\definecolor{OrangeRed4}{rgb}{0.55,0.15,0.00}
\definecolor{OrangeRed}{rgb}{1.00,0.27,0.00}
\definecolor{PaleGoldenrod}{rgb}{0.93,0.91,0.67}
\definecolor{PaleGreen1}{rgb}{0.60,1.00,0.60}
\definecolor{PaleGreen2}{rgb}{0.56,0.93,0.56}
\definecolor{PaleGreen3}{rgb}{0.49,0.80,0.49}
\definecolor{PaleGreen4}{rgb}{0.33,0.55,0.33}
\definecolor{PaleGreen}{rgb}{0.60,0.98,0.60}
\definecolor{PaleTurquoise1}{rgb}{0.73,1.00,1.00}
\definecolor{PaleTurquoise2}{rgb}{0.68,0.93,0.93}
\definecolor{PaleTurquoise3}{rgb}{0.59,0.80,0.80}
\definecolor{PaleTurquoise4}{rgb}{0.40,0.55,0.55}
\definecolor{PaleTurquoise}{rgb}{0.69,0.93,0.93}
\definecolor{PaleVioletRed1}{rgb}{1.00,0.51,0.67}
\definecolor{PaleVioletRed2}{rgb}{0.93,0.47,0.62}
\definecolor{PaleVioletRed3}{rgb}{0.80,0.41,0.54}
\definecolor{PaleVioletRed4}{rgb}{0.55,0.28,0.36}
\definecolor{PaleVioletRed}{rgb}{0.86,0.44,0.58}
\definecolor{PapayaWhip}{rgb}{1.00,0.94,0.84}
\definecolor{PeachPuff1}{rgb}{1.00,0.85,0.73}
\definecolor{PeachPuff2}{rgb}{0.93,0.80,0.68}
\definecolor{PeachPuff3}{rgb}{0.80,0.69,0.58}
\definecolor{PeachPuff4}{rgb}{0.55,0.47,0.40}
\definecolor{PeachPuff}{rgb}{1.00,0.85,0.73}
\definecolor{PowderBlue}{rgb}{0.69,0.88,0.90}
\definecolor{RosyBrown1}{rgb}{1.00,0.76,0.76}
\definecolor{RosyBrown2}{rgb}{0.93,0.71,0.71}
\definecolor{RosyBrown3}{rgb}{0.80,0.61,0.61}
\definecolor{RosyBrown4}{rgb}{0.55,0.41,0.41}
\definecolor{RosyBrown}{rgb}{0.74,0.56,0.56}
\definecolor{RoyalBlue1}{rgb}{0.28,0.46,1.00}
\definecolor{RoyalBlue2}{rgb}{0.26,0.43,0.93}
\definecolor{RoyalBlue3}{rgb}{0.23,0.37,0.80}
\definecolor{RoyalBlue4}{rgb}{0.15,0.25,0.55}
\definecolor{RoyalBlue}{rgb}{0.25,0.41,0.88}
\definecolor{SaddleBrown}{rgb}{0.55,0.27,0.07}
\definecolor{SandyBrown}{rgb}{0.96,0.64,0.38}
\definecolor{SeaGreen1}{rgb}{0.33,1.00,0.62}
\definecolor{SeaGreen2}{rgb}{0.31,0.93,0.58}
\definecolor{SeaGreen3}{rgb}{0.26,0.80,0.50}
\definecolor{SeaGreen4}{rgb}{0.18,0.55,0.34}
\definecolor{SeaGreen}{rgb}{0.18,0.55,0.34}
\definecolor{SkyBlue1}{rgb}{0.53,0.81,1.00}
\definecolor{SkyBlue2}{rgb}{0.49,0.75,0.93}
\definecolor{SkyBlue3}{rgb}{0.42,0.65,0.80}
\definecolor{SkyBlue4}{rgb}{0.29,0.44,0.55}
\definecolor{SkyBlue}{rgb}{0.53,0.81,0.92}
\definecolor{SlateBlue1}{rgb}{0.51,0.44,1.00}
\definecolor{SlateBlue2}{rgb}{0.48,0.40,0.93}
\definecolor{SlateBlue3}{rgb}{0.41,0.35,0.80}
\definecolor{SlateBlue4}{rgb}{0.28,0.24,0.55}
\definecolor{SlateBlue}{rgb}{0.42,0.35,0.80}
\definecolor{SlateGray1}{rgb}{0.78,0.89,1.00}
\definecolor{SlateGray2}{rgb}{0.73,0.83,0.93}
\definecolor{SlateGray3}{rgb}{0.62,0.71,0.80}
\definecolor{SlateGray4}{rgb}{0.42,0.48,0.55}
\definecolor{SlateGray}{rgb}{0.44,0.50,0.56}
\definecolor{SlateGrey}{rgb}{0.44,0.50,0.56}
\definecolor{SpringGreen1}{rgb}{0.00,1.00,0.50}
\definecolor{SpringGreen2}{rgb}{0.00,0.93,0.46}
\definecolor{SpringGreen3}{rgb}{0.00,0.80,0.40}
\definecolor{SpringGreen4}{rgb}{0.00,0.55,0.27}
\definecolor{SpringGreen}{rgb}{0.00,1.00,0.50}
\definecolor{SteelBlue1}{rgb}{0.39,0.72,1.00}
\definecolor{SteelBlue2}{rgb}{0.36,0.67,0.93}
\definecolor{SteelBlue3}{rgb}{0.31,0.58,0.80}
\definecolor{SteelBlue4}{rgb}{0.21,0.39,0.55}
\definecolor{SteelBlue}{rgb}{0.27,0.51,0.71}
\definecolor{VioletRed1}{rgb}{1.00,0.24,0.59}
\definecolor{VioletRed2}{rgb}{0.93,0.23,0.55}
\definecolor{VioletRed3}{rgb}{0.80,0.20,0.47}
\definecolor{VioletRed4}{rgb}{0.55,0.13,0.32}
\definecolor{VioletRed}{rgb}{0.82,0.13,0.56}
\definecolor{WhiteSmoke}{rgb}{0.96,0.96,0.96}
\definecolor{YellowGreen}{rgb}{0.60,0.80,0.20}
\definecolor{aliceblue}{rgb}{0.94,0.97,1.00}
\definecolor{antiquewhite}{rgb}{0.98,0.92,0.84}
\definecolor{aquamarine1}{rgb}{0.50,1.00,0.83}
\definecolor{aquamarine2}{rgb}{0.46,0.93,0.78}
\definecolor{aquamarine3}{rgb}{0.40,0.80,0.67}
\definecolor{aquamarine4}{rgb}{0.27,0.55,0.45}
\definecolor{aquamarine}{rgb}{0.50,1.00,0.83}
\definecolor{azure1}{rgb}{0.94,1.00,1.00}
\definecolor{azure2}{rgb}{0.88,0.93,0.93}
\definecolor{azure3}{rgb}{0.76,0.80,0.80}
\definecolor{azure4}{rgb}{0.51,0.55,0.55}
\definecolor{azure}{rgb}{0.94,1.00,1.00}
\definecolor{beige}{rgb}{0.96,0.96,0.86}
\definecolor{bisque1}{rgb}{1.00,0.89,0.77}
\definecolor{bisque2}{rgb}{0.93,0.84,0.72}
\definecolor{bisque3}{rgb}{0.80,0.72,0.62}
\definecolor{bisque4}{rgb}{0.55,0.49,0.42}
\definecolor{bisque}{rgb}{1.00,0.89,0.77}
\definecolor{black}{rgb}{0.00,0.00,0.00}
\definecolor{blanchedalmond}{rgb}{1.00,0.92,0.80}
\definecolor{blue1}{rgb}{0.00,0.00,1.00}
\definecolor{blue2}{rgb}{0.00,0.00,0.93}
\definecolor{blue3}{rgb}{0.00,0.00,0.80}
\definecolor{blue4}{rgb}{0.00,0.00,0.55}
\definecolor{blueviolet}{rgb}{0.54,0.17,0.89}
\definecolor{blue}{rgb}{0.00,0.00,1.00}
\definecolor{brown1}{rgb}{1.00,0.25,0.25}
\definecolor{brown2}{rgb}{0.93,0.23,0.23}
\definecolor{brown3}{rgb}{0.80,0.20,0.20}
\definecolor{brown4}{rgb}{0.55,0.14,0.14}
\definecolor{brown}{rgb}{0.65,0.16,0.16}
\definecolor{burlywood1}{rgb}{1.00,0.83,0.61}
\definecolor{burlywood2}{rgb}{0.93,0.77,0.57}
\definecolor{burlywood3}{rgb}{0.80,0.67,0.49}
\definecolor{burlywood4}{rgb}{0.55,0.45,0.33}
\definecolor{burlywood}{rgb}{0.87,0.72,0.53}
\definecolor{cadetblue}{rgb}{0.37,0.62,0.63}
\definecolor{chartreuse1}{rgb}{0.50,1.00,0.00}
\definecolor{chartreuse2}{rgb}{0.46,0.93,0.00}
\definecolor{chartreuse3}{rgb}{0.40,0.80,0.00}
\definecolor{chartreuse4}{rgb}{0.27,0.55,0.00}
\definecolor{chartreuse}{rgb}{0.50,1.00,0.00}
\definecolor{chocolate1}{rgb}{1.00,0.50,0.14}
\definecolor{chocolate2}{rgb}{0.93,0.46,0.13}
\definecolor{chocolate3}{rgb}{0.80,0.40,0.11}
\definecolor{chocolate4}{rgb}{0.55,0.27,0.07}
\definecolor{chocolate}{rgb}{0.82,0.41,0.12}
\definecolor{coral1}{rgb}{1.00,0.45,0.34}
\definecolor{coral2}{rgb}{0.93,0.42,0.31}
\definecolor{coral3}{rgb}{0.80,0.36,0.27}
\definecolor{coral4}{rgb}{0.55,0.24,0.18}
\definecolor{coral}{rgb}{1.00,0.50,0.31}
\definecolor{cornflowerblue}{rgb}{0.39,0.58,0.93}
\definecolor{cornsilk1}{rgb}{1.00,0.97,0.86}
\definecolor{cornsilk2}{rgb}{0.93,0.91,0.80}
\definecolor{cornsilk3}{rgb}{0.80,0.78,0.69}
\definecolor{cornsilk4}{rgb}{0.55,0.53,0.47}
\definecolor{cornsilk}{rgb}{1.00,0.97,0.86}
\definecolor{cyan1}{rgb}{0.00,1.00,1.00}
\definecolor{cyan2}{rgb}{0.00,0.93,0.93}
\definecolor{cyan3}{rgb}{0.00,0.80,0.80}
\definecolor{cyan4}{rgb}{0.00,0.55,0.55}
\definecolor{cyan}{rgb}{0.00,1.00,1.00}
\definecolor{darkblue}{rgb}{0.00,0.00,0.55}
\definecolor{darkcyan}{rgb}{0.00,0.55,0.55}
\definecolor{darkgoldenrod}{rgb}{0.72,0.53,0.04}
\definecolor{darkgray}{rgb}{0.66,0.66,0.66}
\definecolor{darkgreen}{rgb}{0.00,0.39,0.00}
\definecolor{darkgrey}{rgb}{0.66,0.66,0.66}
\definecolor{darkkhaki}{rgb}{0.74,0.72,0.42}
\definecolor{darkmagenta}{rgb}{0.55,0.00,0.55}
\definecolor{darkolive}{rgb}{0.33,0.42,0.18}
\definecolor{darkorange}{rgb}{1.00,0.55,0.00}
\definecolor{darkorchid}{rgb}{0.60,0.20,0.80}
\definecolor{darkred}{rgb}{0.55,0.00,0.00}
\definecolor{darksalmon}{rgb}{0.91,0.59,0.48}
\definecolor{darksea}{rgb}{0.56,0.74,0.56}
\definecolor{darkslate}{rgb}{0.18,0.31,0.31}
\definecolor{darkslate}{rgb}{0.18,0.31,0.31}
\definecolor{darkslate}{rgb}{0.28,0.24,0.55}
\definecolor{darkturquoise}{rgb}{0.00,0.81,0.82}
\definecolor{darkviolet}{rgb}{0.58,0.00,0.83}
\definecolor{deeppink}{rgb}{1.00,0.08,0.58}
\definecolor{deepsky}{rgb}{0.00,0.75,1.00}
\definecolor{dimgray}{rgb}{0.41,0.41,0.41}
\definecolor{dimgrey}{rgb}{0.41,0.41,0.41}
\definecolor{dodgerblue}{rgb}{0.12,0.56,1.00}
\definecolor{firebrick1}{rgb}{1.00,0.19,0.19}
\definecolor{firebrick2}{rgb}{0.93,0.17,0.17}
\definecolor{firebrick3}{rgb}{0.80,0.15,0.15}
\definecolor{firebrick4}{rgb}{0.55,0.10,0.10}
\definecolor{firebrick}{rgb}{0.70,0.13,0.13}
\definecolor{floralwhite}{rgb}{1.00,0.98,0.94}
\definecolor{forestgreen}{rgb}{0.13,0.55,0.13}
\definecolor{gainsboro}{rgb}{0.86,0.86,0.86}
\definecolor{ghostwhite}{rgb}{0.97,0.97,1.00}
\definecolor{gold1}{rgb}{1.00,0.84,0.00}
\definecolor{gold2}{rgb}{0.93,0.79,0.00}
\definecolor{gold3}{rgb}{0.80,0.68,0.00}
\definecolor{gold4}{rgb}{0.55,0.46,0.00}
\definecolor{goldenrod1}{rgb}{1.00,0.76,0.15}
\definecolor{goldenrod2}{rgb}{0.93,0.71,0.13}
\definecolor{goldenrod3}{rgb}{0.80,0.61,0.11}
\definecolor{goldenrod4}{rgb}{0.55,0.41,0.08}
\definecolor{goldenrod}{rgb}{0.85,0.65,0.13}
\definecolor{gold}{rgb}{1.00,0.84,0.00}
\definecolor{gray0}{rgb}{0.00,0.00,0.00}
\definecolor{gray100}{rgb}{1.00,1.00,1.00}
\definecolor{gray10}{rgb}{0.10,0.10,0.10}
\definecolor{gray11}{rgb}{0.11,0.11,0.11}
\definecolor{gray12}{rgb}{0.12,0.12,0.12}
\definecolor{gray13}{rgb}{0.13,0.13,0.13}
\definecolor{gray14}{rgb}{0.14,0.14,0.14}
\definecolor{gray15}{rgb}{0.15,0.15,0.15}
\definecolor{gray16}{rgb}{0.16,0.16,0.16}
\definecolor{gray17}{rgb}{0.17,0.17,0.17}
\definecolor{gray18}{rgb}{0.18,0.18,0.18}
\definecolor{gray19}{rgb}{0.19,0.19,0.19}
\definecolor{gray1}{rgb}{0.01,0.01,0.01}
\definecolor{gray20}{rgb}{0.20,0.20,0.20}
\definecolor{gray21}{rgb}{0.21,0.21,0.21}
\definecolor{gray22}{rgb}{0.22,0.22,0.22}
\definecolor{gray23}{rgb}{0.23,0.23,0.23}
\definecolor{gray24}{rgb}{0.24,0.24,0.24}
\definecolor{gray25}{rgb}{0.25,0.25,0.25}
\definecolor{gray26}{rgb}{0.26,0.26,0.26}
\definecolor{gray27}{rgb}{0.27,0.27,0.27}
\definecolor{gray28}{rgb}{0.28,0.28,0.28}
\definecolor{gray29}{rgb}{0.29,0.29,0.29}
\definecolor{gray2}{rgb}{0.02,0.02,0.02}
\definecolor{gray30}{rgb}{0.30,0.30,0.30}
\definecolor{gray31}{rgb}{0.31,0.31,0.31}
\definecolor{gray32}{rgb}{0.32,0.32,0.32}
\definecolor{gray33}{rgb}{0.33,0.33,0.33}
\definecolor{gray34}{rgb}{0.34,0.34,0.34}
\definecolor{gray35}{rgb}{0.35,0.35,0.35}
\definecolor{gray36}{rgb}{0.36,0.36,0.36}
\definecolor{gray37}{rgb}{0.37,0.37,0.37}
\definecolor{gray38}{rgb}{0.38,0.38,0.38}
\definecolor{gray39}{rgb}{0.39,0.39,0.39}
\definecolor{gray3}{rgb}{0.03,0.03,0.03}
\definecolor{gray40}{rgb}{0.40,0.40,0.40}
\definecolor{gray41}{rgb}{0.41,0.41,0.41}
\definecolor{gray42}{rgb}{0.42,0.42,0.42}
\definecolor{gray43}{rgb}{0.43,0.43,0.43}
\definecolor{gray44}{rgb}{0.44,0.44,0.44}
\definecolor{gray45}{rgb}{0.45,0.45,0.45}
\definecolor{gray46}{rgb}{0.46,0.46,0.46}
\definecolor{gray47}{rgb}{0.47,0.47,0.47}
\definecolor{gray48}{rgb}{0.48,0.48,0.48}
\definecolor{gray49}{rgb}{0.49,0.49,0.49}
\definecolor{gray4}{rgb}{0.04,0.04,0.04}
\definecolor{gray50}{rgb}{0.50,0.50,0.50}
\definecolor{gray51}{rgb}{0.51,0.51,0.51}
\definecolor{gray52}{rgb}{0.52,0.52,0.52}
\definecolor{gray53}{rgb}{0.53,0.53,0.53}
\definecolor{gray54}{rgb}{0.54,0.54,0.54}
\definecolor{gray55}{rgb}{0.55,0.55,0.55}
\definecolor{gray56}{rgb}{0.56,0.56,0.56}
\definecolor{gray57}{rgb}{0.57,0.57,0.57}
\definecolor{gray58}{rgb}{0.58,0.58,0.58}
\definecolor{gray59}{rgb}{0.59,0.59,0.59}
\definecolor{gray5}{rgb}{0.05,0.05,0.05}
\definecolor{gray60}{rgb}{0.60,0.60,0.60}
\definecolor{gray61}{rgb}{0.61,0.61,0.61}
\definecolor{gray62}{rgb}{0.62,0.62,0.62}
\definecolor{gray63}{rgb}{0.63,0.63,0.63}
\definecolor{gray64}{rgb}{0.64,0.64,0.64}
\definecolor{gray65}{rgb}{0.65,0.65,0.65}
\definecolor{gray66}{rgb}{0.66,0.66,0.66}
\definecolor{gray67}{rgb}{0.67,0.67,0.67}
\definecolor{gray68}{rgb}{0.68,0.68,0.68}
\definecolor{gray69}{rgb}{0.69,0.69,0.69}
\definecolor{gray6}{rgb}{0.06,0.06,0.06}
\definecolor{gray70}{rgb}{0.70,0.70,0.70}
\definecolor{gray71}{rgb}{0.71,0.71,0.71}
\definecolor{gray72}{rgb}{0.72,0.72,0.72}
\definecolor{gray73}{rgb}{0.73,0.73,0.73}
\definecolor{gray74}{rgb}{0.74,0.74,0.74}
\definecolor{gray75}{rgb}{0.75,0.75,0.75}
\definecolor{gray76}{rgb}{0.76,0.76,0.76}
\definecolor{gray77}{rgb}{0.77,0.77,0.77}
\definecolor{gray78}{rgb}{0.78,0.78,0.78}
\definecolor{gray79}{rgb}{0.79,0.79,0.79}
\definecolor{gray7}{rgb}{0.07,0.07,0.07}
\definecolor{gray80}{rgb}{0.80,0.80,0.80}
\definecolor{gray81}{rgb}{0.81,0.81,0.81}
\definecolor{gray82}{rgb}{0.82,0.82,0.82}
\definecolor{gray83}{rgb}{0.83,0.83,0.83}
\definecolor{gray84}{rgb}{0.84,0.84,0.84}
\definecolor{gray85}{rgb}{0.85,0.85,0.85}
\definecolor{gray86}{rgb}{0.86,0.86,0.86}
\definecolor{gray87}{rgb}{0.87,0.87,0.87}
\definecolor{gray88}{rgb}{0.88,0.88,0.88}
\definecolor{gray89}{rgb}{0.89,0.89,0.89}
\definecolor{gray8}{rgb}{0.08,0.08,0.08}
\definecolor{gray90}{rgb}{0.90,0.90,0.90}
\definecolor{gray91}{rgb}{0.91,0.91,0.91}
\definecolor{gray92}{rgb}{0.92,0.92,0.92}
\definecolor{gray93}{rgb}{0.93,0.93,0.93}
\definecolor{gray94}{rgb}{0.94,0.94,0.94}
\definecolor{gray95}{rgb}{0.95,0.95,0.95}
\definecolor{gray96}{rgb}{0.96,0.96,0.96}
\definecolor{gray97}{rgb}{0.97,0.97,0.97}
\definecolor{gray98}{rgb}{0.98,0.98,0.98}
\definecolor{gray99}{rgb}{0.99,0.99,0.99}
\definecolor{gray9}{rgb}{0.09,0.09,0.09}
\definecolor{gray}{rgb}{0.75,0.75,0.75}
\definecolor{green1}{rgb}{0.00,1.00,0.00}
\definecolor{green2}{rgb}{0.00,0.93,0.00}
\definecolor{green3}{rgb}{0.00,0.80,0.00}
\definecolor{green4}{rgb}{0.00,0.55,0.00}
\definecolor{greenyellow}{rgb}{0.68,1.00,0.18}
\definecolor{green}{rgb}{0.00,1.00,0.00}
\definecolor{grey0}{rgb}{0.00,0.00,0.00}
\definecolor{grey100}{rgb}{1.00,1.00,1.00}
\definecolor{grey10}{rgb}{0.10,0.10,0.10}
\definecolor{grey11}{rgb}{0.11,0.11,0.11}
\definecolor{grey12}{rgb}{0.12,0.12,0.12}
\definecolor{grey13}{rgb}{0.13,0.13,0.13}
\definecolor{grey14}{rgb}{0.14,0.14,0.14}
\definecolor{grey15}{rgb}{0.15,0.15,0.15}
\definecolor{grey16}{rgb}{0.16,0.16,0.16}
\definecolor{grey17}{rgb}{0.17,0.17,0.17}
\definecolor{grey18}{rgb}{0.18,0.18,0.18}
\definecolor{grey19}{rgb}{0.19,0.19,0.19}
\definecolor{grey1}{rgb}{0.01,0.01,0.01}
\definecolor{grey20}{rgb}{0.20,0.20,0.20}
\definecolor{grey21}{rgb}{0.21,0.21,0.21}
\definecolor{grey22}{rgb}{0.22,0.22,0.22}
\definecolor{grey23}{rgb}{0.23,0.23,0.23}
\definecolor{grey24}{rgb}{0.24,0.24,0.24}
\definecolor{grey25}{rgb}{0.25,0.25,0.25}
\definecolor{grey26}{rgb}{0.26,0.26,0.26}
\definecolor{grey27}{rgb}{0.27,0.27,0.27}
\definecolor{grey28}{rgb}{0.28,0.28,0.28}
\definecolor{grey29}{rgb}{0.29,0.29,0.29}
\definecolor{grey2}{rgb}{0.02,0.02,0.02}
\definecolor{grey30}{rgb}{0.30,0.30,0.30}
\definecolor{grey31}{rgb}{0.31,0.31,0.31}
\definecolor{grey32}{rgb}{0.32,0.32,0.32}
\definecolor{grey33}{rgb}{0.33,0.33,0.33}
\definecolor{grey34}{rgb}{0.34,0.34,0.34}
\definecolor{grey35}{rgb}{0.35,0.35,0.35}
\definecolor{grey36}{rgb}{0.36,0.36,0.36}
\definecolor{grey37}{rgb}{0.37,0.37,0.37}
\definecolor{grey38}{rgb}{0.38,0.38,0.38}
\definecolor{grey39}{rgb}{0.39,0.39,0.39}
\definecolor{grey3}{rgb}{0.03,0.03,0.03}
\definecolor{grey40}{rgb}{0.40,0.40,0.40}
\definecolor{grey41}{rgb}{0.41,0.41,0.41}
\definecolor{grey42}{rgb}{0.42,0.42,0.42}
\definecolor{grey43}{rgb}{0.43,0.43,0.43}
\definecolor{grey44}{rgb}{0.44,0.44,0.44}
\definecolor{grey45}{rgb}{0.45,0.45,0.45}
\definecolor{grey46}{rgb}{0.46,0.46,0.46}
\definecolor{grey47}{rgb}{0.47,0.47,0.47}
\definecolor{grey48}{rgb}{0.48,0.48,0.48}
\definecolor{grey49}{rgb}{0.49,0.49,0.49}
\definecolor{grey4}{rgb}{0.04,0.04,0.04}
\definecolor{grey50}{rgb}{0.50,0.50,0.50}
\definecolor{grey51}{rgb}{0.51,0.51,0.51}
\definecolor{grey52}{rgb}{0.52,0.52,0.52}
\definecolor{grey53}{rgb}{0.53,0.53,0.53}
\definecolor{grey54}{rgb}{0.54,0.54,0.54}
\definecolor{grey55}{rgb}{0.55,0.55,0.55}
\definecolor{grey56}{rgb}{0.56,0.56,0.56}
\definecolor{grey57}{rgb}{0.57,0.57,0.57}
\definecolor{grey58}{rgb}{0.58,0.58,0.58}
\definecolor{grey59}{rgb}{0.59,0.59,0.59}
\definecolor{grey5}{rgb}{0.05,0.05,0.05}
\definecolor{grey60}{rgb}{0.60,0.60,0.60}
\definecolor{grey61}{rgb}{0.61,0.61,0.61}
\definecolor{grey62}{rgb}{0.62,0.62,0.62}
\definecolor{grey63}{rgb}{0.63,0.63,0.63}
\definecolor{grey64}{rgb}{0.64,0.64,0.64}
\definecolor{grey65}{rgb}{0.65,0.65,0.65}
\definecolor{grey66}{rgb}{0.66,0.66,0.66}
\definecolor{grey67}{rgb}{0.67,0.67,0.67}
\definecolor{grey68}{rgb}{0.68,0.68,0.68}
\definecolor{grey69}{rgb}{0.69,0.69,0.69}
\definecolor{grey6}{rgb}{0.06,0.06,0.06}
\definecolor{grey70}{rgb}{0.70,0.70,0.70}
\definecolor{grey71}{rgb}{0.71,0.71,0.71}
\definecolor{grey72}{rgb}{0.72,0.72,0.72}
\definecolor{grey73}{rgb}{0.73,0.73,0.73}
\definecolor{grey74}{rgb}{0.74,0.74,0.74}
\definecolor{grey75}{rgb}{0.75,0.75,0.75}
\definecolor{grey76}{rgb}{0.76,0.76,0.76}
\definecolor{grey77}{rgb}{0.77,0.77,0.77}
\definecolor{grey78}{rgb}{0.78,0.78,0.78}
\definecolor{grey79}{rgb}{0.79,0.79,0.79}
\definecolor{grey7}{rgb}{0.07,0.07,0.07}
\definecolor{grey80}{rgb}{0.80,0.80,0.80}
\definecolor{grey81}{rgb}{0.81,0.81,0.81}
\definecolor{grey82}{rgb}{0.82,0.82,0.82}
\definecolor{grey83}{rgb}{0.83,0.83,0.83}
\definecolor{grey84}{rgb}{0.84,0.84,0.84}
\definecolor{grey85}{rgb}{0.85,0.85,0.85}
\definecolor{grey86}{rgb}{0.86,0.86,0.86}
\definecolor{grey87}{rgb}{0.87,0.87,0.87}
\definecolor{grey88}{rgb}{0.88,0.88,0.88}
\definecolor{grey89}{rgb}{0.89,0.89,0.89}
\definecolor{grey8}{rgb}{0.08,0.08,0.08}
\definecolor{grey90}{rgb}{0.90,0.90,0.90}
\definecolor{grey91}{rgb}{0.91,0.91,0.91}
\definecolor{grey92}{rgb}{0.92,0.92,0.92}
\definecolor{grey93}{rgb}{0.93,0.93,0.93}
\definecolor{grey94}{rgb}{0.94,0.94,0.94}
\definecolor{grey95}{rgb}{0.95,0.95,0.95}
\definecolor{grey96}{rgb}{0.96,0.96,0.96}
\definecolor{grey97}{rgb}{0.97,0.97,0.97}
\definecolor{grey98}{rgb}{0.98,0.98,0.98}
\definecolor{grey99}{rgb}{0.99,0.99,0.99}
\definecolor{grey9}{rgb}{0.09,0.09,0.09}
\definecolor{grey}{rgb}{0.75,0.75,0.75}
\definecolor{honeydew1}{rgb}{0.94,1.00,0.94}
\definecolor{honeydew2}{rgb}{0.88,0.93,0.88}
\definecolor{honeydew3}{rgb}{0.76,0.80,0.76}
\definecolor{honeydew4}{rgb}{0.51,0.55,0.51}
\definecolor{honeydew}{rgb}{0.94,1.00,0.94}
\definecolor{hotpink}{rgb}{1.00,0.41,0.71}
\definecolor{indianred}{rgb}{0.80,0.36,0.36}
\definecolor{ivory1}{rgb}{1.00,1.00,0.94}
\definecolor{ivory2}{rgb}{0.93,0.93,0.88}
\definecolor{ivory3}{rgb}{0.80,0.80,0.76}
\definecolor{ivory4}{rgb}{0.55,0.55,0.51}
\definecolor{ivory}{rgb}{1.00,1.00,0.94}
\definecolor{khaki1}{rgb}{1.00,0.96,0.56}
\definecolor{khaki2}{rgb}{0.93,0.90,0.52}
\definecolor{khaki3}{rgb}{0.80,0.78,0.45}
\definecolor{khaki4}{rgb}{0.55,0.53,0.31}
\definecolor{khaki}{rgb}{0.94,0.90,0.55}
\definecolor{lavenderblush}{rgb}{1.00,0.94,0.96}
\definecolor{lavender}{rgb}{0.90,0.90,0.98}
\definecolor{lawngreen}{rgb}{0.49,0.99,0.00}
\definecolor{lemonchiffon}{rgb}{1.00,0.98,0.80}
\definecolor{lightblue}{rgb}{0.68,0.85,0.90}
\definecolor{lightcoral}{rgb}{0.94,0.50,0.50}
\definecolor{lightcyan}{rgb}{0.88,1.00,1.00}
\definecolor{lightgoldenrod}{rgb}{0.93,0.87,0.51}
\definecolor{lightgoldenrod}{rgb}{0.98,0.98,0.82}
\definecolor{lightgray}{rgb}{0.83,0.83,0.83}
\definecolor{lightgreen}{rgb}{0.56,0.93,0.56}
\definecolor{lightgrey}{rgb}{0.83,0.83,0.83}
\definecolor{lightpink}{rgb}{1.00,0.71,0.76}
\definecolor{lightsalmon}{rgb}{1.00,0.63,0.48}
\definecolor{lightsea}{rgb}{0.13,0.70,0.67}
\definecolor{lightsky}{rgb}{0.53,0.81,0.98}
\definecolor{lightslate}{rgb}{0.47,0.53,0.60}
\definecolor{lightslate}{rgb}{0.47,0.53,0.60}
\definecolor{lightslate}{rgb}{0.52,0.44,1.00}
\definecolor{lightsteel}{rgb}{0.69,0.77,0.87}
\definecolor{lightyellow}{rgb}{1.00,1.00,0.88}
\definecolor{limegreen}{rgb}{0.20,0.80,0.20}
\definecolor{linen}{rgb}{0.98,0.94,0.90}
\definecolor{magenta1}{rgb}{1.00,0.00,1.00}
\definecolor{magenta2}{rgb}{0.93,0.00,0.93}
\definecolor{magenta3}{rgb}{0.80,0.00,0.80}
\definecolor{magenta4}{rgb}{0.55,0.00,0.55}
\definecolor{magenta}{rgb}{1.00,0.00,1.00}
\definecolor{maroon1}{rgb}{1.00,0.20,0.70}
\definecolor{maroon2}{rgb}{0.93,0.19,0.65}
\definecolor{maroon3}{rgb}{0.80,0.16,0.56}
\definecolor{maroon4}{rgb}{0.55,0.11,0.38}
\definecolor{maroon}{rgb}{0.69,0.19,0.38}
\definecolor{mediumaquamarine}{rgb}{0.40,0.80,0.67}
\definecolor{mediumblue}{rgb}{0.00,0.00,0.80}
\definecolor{mediumorchid}{rgb}{0.73,0.33,0.83}
\definecolor{mediumpurple}{rgb}{0.58,0.44,0.86}
\definecolor{mediumsea}{rgb}{0.24,0.70,0.44}
\definecolor{mediumslate}{rgb}{0.48,0.41,0.93}
\definecolor{mediumspring}{rgb}{0.00,0.98,0.60}
\definecolor{mediumturquoise}{rgb}{0.28,0.82,0.80}
\definecolor{mediumviolet}{rgb}{0.78,0.08,0.52}
\definecolor{midnightblue}{rgb}{0.10,0.10,0.44}
\definecolor{mintcream}{rgb}{0.96,1.00,0.98}
\definecolor{mistyrose}{rgb}{1.00,0.89,0.88}
\definecolor{moccasin}{rgb}{1.00,0.89,0.71}
\definecolor{navajowhite}{rgb}{1.00,0.87,0.68}
\definecolor{navyblue}{rgb}{0.00,0.00,0.50}
\definecolor{navy}{rgb}{0.00,0.00,0.50}
\definecolor{oldlace}{rgb}{0.99,0.96,0.90}
\definecolor{olivedrab}{rgb}{0.42,0.56,0.14}
\definecolor{orange1}{rgb}{1.00,0.65,0.00}
\definecolor{orange2}{rgb}{0.93,0.60,0.00}
\definecolor{orange3}{rgb}{0.80,0.52,0.00}
\definecolor{orange4}{rgb}{0.55,0.35,0.00}
\definecolor{orangered}{rgb}{1.00,0.27,0.00}
\definecolor{orange}{rgb}{1.00,0.65,0.00}
\definecolor{orchid1}{rgb}{1.00,0.51,0.98}
\definecolor{orchid2}{rgb}{0.93,0.48,0.91}
\definecolor{orchid3}{rgb}{0.80,0.41,0.79}
\definecolor{orchid4}{rgb}{0.55,0.28,0.54}
\definecolor{orchid}{rgb}{0.85,0.44,0.84}
\definecolor{palegoldenrod}{rgb}{0.93,0.91,0.67}
\definecolor{palegreen}{rgb}{0.60,0.98,0.60}
\definecolor{paleturquoise}{rgb}{0.69,0.93,0.93}
\definecolor{paleviolet}{rgb}{0.86,0.44,0.58}
\definecolor{papayawhip}{rgb}{1.00,0.94,0.84}
\definecolor{peachpuff}{rgb}{1.00,0.85,0.73}
\definecolor{peru}{rgb}{0.80,0.52,0.25}
\definecolor{pink1}{rgb}{1.00,0.71,0.77}
\definecolor{pink2}{rgb}{0.93,0.66,0.72}
\definecolor{pink3}{rgb}{0.80,0.57,0.62}
\definecolor{pink4}{rgb}{0.55,0.39,0.42}
\definecolor{pink}{rgb}{1.00,0.75,0.80}
\definecolor{plum1}{rgb}{1.00,0.73,1.00}
\definecolor{plum2}{rgb}{0.93,0.68,0.93}
\definecolor{plum3}{rgb}{0.80,0.59,0.80}
\definecolor{plum4}{rgb}{0.55,0.40,0.55}
\definecolor{plum}{rgb}{0.87,0.63,0.87}
\definecolor{powderblue}{rgb}{0.69,0.88,0.90}
\definecolor{purple1}{rgb}{0.61,0.19,1.00}
\definecolor{purple2}{rgb}{0.57,0.17,0.93}
\definecolor{purple3}{rgb}{0.49,0.15,0.80}
\definecolor{purple4}{rgb}{0.33,0.10,0.55}
\definecolor{purple}{rgb}{0.63,0.13,0.94}
\definecolor{red1}{rgb}{1.00,0.00,0.00}
\definecolor{red2}{rgb}{0.93,0.00,0.00}
\definecolor{red3}{rgb}{0.80,0.00,0.00}
\definecolor{red4}{rgb}{0.55,0.00,0.00}
\definecolor{red}{rgb}{1.00,0.00,0.00}
\definecolor{rosybrown}{rgb}{0.74,0.56,0.56}
\definecolor{royalblue}{rgb}{0.25,0.41,0.88}
\definecolor{saddlebrown}{rgb}{0.55,0.27,0.07}
\definecolor{salmon1}{rgb}{1.00,0.55,0.41}
\definecolor{salmon2}{rgb}{0.93,0.51,0.38}
\definecolor{salmon3}{rgb}{0.80,0.44,0.33}
\definecolor{salmon4}{rgb}{0.55,0.30,0.22}
\definecolor{salmon}{rgb}{0.98,0.50,0.45}
\definecolor{sandybrown}{rgb}{0.96,0.64,0.38}
\definecolor{seagreen}{rgb}{0.18,0.55,0.34}
\definecolor{seashell1}{rgb}{1.00,0.96,0.93}
\definecolor{seashell2}{rgb}{0.93,0.90,0.87}
\definecolor{seashell3}{rgb}{0.80,0.77,0.75}
\definecolor{seashell4}{rgb}{0.55,0.53,0.51}
\definecolor{seashell}{rgb}{1.00,0.96,0.93}
\definecolor{sienna1}{rgb}{1.00,0.51,0.28}
\definecolor{sienna2}{rgb}{0.93,0.47,0.26}
\definecolor{sienna3}{rgb}{0.80,0.41,0.22}
\definecolor{sienna4}{rgb}{0.55,0.28,0.15}
\definecolor{sienna}{rgb}{0.63,0.32,0.18}
\definecolor{skyblue}{rgb}{0.53,0.81,0.92}
\definecolor{slateblue}{rgb}{0.42,0.35,0.80}
\definecolor{slategray}{rgb}{0.44,0.50,0.56}
\definecolor{slategrey}{rgb}{0.44,0.50,0.56}
\definecolor{snow1}{rgb}{1.00,0.98,0.98}
\definecolor{snow2}{rgb}{0.93,0.91,0.91}
\definecolor{snow3}{rgb}{0.80,0.79,0.79}
\definecolor{snow4}{rgb}{0.55,0.54,0.54}
\definecolor{snow}{rgb}{1.00,0.98,0.98}
\definecolor{springgreen}{rgb}{0.00,1.00,0.50}
\definecolor{steelblue}{rgb}{0.27,0.51,0.71}
\definecolor{tan1}{rgb}{1.00,0.65,0.31}
\definecolor{tan2}{rgb}{0.93,0.60,0.29}
\definecolor{tan3}{rgb}{0.80,0.52,0.25}
\definecolor{tan4}{rgb}{0.55,0.35,0.17}
\definecolor{tan}{rgb}{0.82,0.71,0.55}
\definecolor{thistle1}{rgb}{1.00,0.88,1.00}
\definecolor{thistle2}{rgb}{0.93,0.82,0.93}
\definecolor{thistle3}{rgb}{0.80,0.71,0.80}
\definecolor{thistle4}{rgb}{0.55,0.48,0.55}
\definecolor{thistle}{rgb}{0.85,0.75,0.85}
\definecolor{tomato1}{rgb}{1.00,0.39,0.28}
\definecolor{tomato2}{rgb}{0.93,0.36,0.26}
\definecolor{tomato3}{rgb}{0.80,0.31,0.22}
\definecolor{tomato4}{rgb}{0.55,0.21,0.15}
\definecolor{tomato}{rgb}{1.00,0.39,0.28}
\definecolor{turquoise1}{rgb}{0.00,0.96,1.00}
\definecolor{turquoise2}{rgb}{0.00,0.90,0.93}
\definecolor{turquoise3}{rgb}{0.00,0.77,0.80}
\definecolor{turquoise4}{rgb}{0.00,0.53,0.55}
\definecolor{turquoise}{rgb}{0.25,0.88,0.82}
\definecolor{violetred}{rgb}{0.82,0.13,0.56}
\definecolor{violet}{rgb}{0.93,0.51,0.93}
\definecolor{wheat1}{rgb}{1.00,0.91,0.73}
\definecolor{wheat2}{rgb}{0.93,0.85,0.68}
\definecolor{wheat3}{rgb}{0.80,0.73,0.59}
\definecolor{wheat4}{rgb}{0.55,0.49,0.40}
\definecolor{wheat}{rgb}{0.96,0.87,0.70}
\definecolor{whitesmoke}{rgb}{0.96,0.96,0.96}
\definecolor{white}{rgb}{1.00,1.00,1.00}
\definecolor{yellow1}{rgb}{1.00,1.00,0.00}
\definecolor{yellow2}{rgb}{0.93,0.93,0.00}
\definecolor{yellow3}{rgb}{0.80,0.80,0.00}
\definecolor{yellow4}{rgb}{0.55,0.55,0.00}
\definecolor{yellowgreen}{rgb}{0.60,0.80,0.20}
\definecolor{yellow}{rgb}{1.00,1.00,0.00}

\usepackage{subfigure} 

\usepackage{url}       

\usepackage{stfloats}  
\usepackage{amssymb}
\usepackage{amsfonts}
\usepackage{amsmath}   
\usepackage{pifont}

\usepackage{algorithmicx}
\usepackage{algorithm}
\usepackage{algpseudocode}

\begin{document}
%
\title{Robust Registration of Gaussian Mixtures for Colour Transfer}
%
%
\author{Mair\'ead Grogan \& Rozenn Dahyot\\
School of Computer Science and Statistics\\
Trinity College Dublin, Ireland\\  
}

\maketitle

\begin{abstract} We present a flexible approach to colour transfer inspired by techniques recently proposed for shape registration. Colour distributions of the
palette and target images  are modelled with Gaussian
Mixture Models (GMMs)  that are robustly registered  to infer  a non linear parametric transfer function. We show experimentally that our approach compares well to current techniques  both quantitatively and qualitatively. 
Moreover, our technique is  computationally  the fastest and can take efficient  advantage of parallel processing architectures for recolouring  images and videos. 
Our transfer function  is parametric and hence can be stored in memory
for later usage and also combined  with other  computed transfer
functions to create interesting visual effects.  Overall  this paper provides  a fast user friendly  approach to recolouring of image and video materials. 
\end{abstract}

\section{Introduction}

Colour transfer refers to a set of techniques that aim
to modify the colour feel of a target image or video using
an exemplar colour palette provided by another image or
video. Most  techniques are based on the idea of warping some colour statistics 
from the target image colour distribution to the palette image colour distribution. 
The transfer (or warping) function $\phi$, once estimated, 
is then used to recolour a colour pixel value $x$ to $\phi(x)$.
\textcolor{black}{We have recently proposed to formulate the colour transfer problem as a shape registration one, whereabout a parametric warping function  $\phi_{\theta}(x)$, controlled by a latent vector $\theta$ of parameters, is directly estimated by minimising the Euclidean $\mathcal{L}_2$ distance between the target and palette probability density functions  capturing the colour content of the palette and target images \cite{Grogan2015,Grogan2015b}. 
This paper proposes several additional improvements. 
First, our  framework is extended so that it can  take advantage of correspondences that may be available   between pixels of the target and palette image. Indeed, when considering target and palette images of the same scene, correspondences can easily be computed  using  registration techniques,
potentially  creating some outlier pairs, and our technique is shown
to be robust to these occurrences. In a more general setting where the content in the palette and target images is different,  user defined correspondences can also be defined to constrain the recolouring, providing some needed controls to artists manipulating images.
Secondly, we explore several clustering techniques when modelling the probability density functions of the target and palette image, as well as several parametric models for the warping function $\phi_{\theta}$.
Finally, recolouring using the estimated warping function $\hat{\phi}=\phi_{\hat{\theta}}$ is implemented using parallel programming techniques, and our approach is shown to currently be the fastest for recolouring. Moreover, our framework allows the user to easily mix several warping functions to create new ones, suitable for generating various spatio-temporal effects for recolouring images and videos.}
First we present a short review of relevant and recent colour
transfer techniques (Section \ref{sec:soa}), as well as techniques for shape
registration that serve as a background for our colour transfer technique ( Section \ref{sec:our:approach})
which combines all the advantages of past methods while providing a computationally efficient and convenient recolouring tool for users.
An exhaustive set of  experiments has been carried out to assess performance against leading techniques in the field (Section \ref{sec:quant:results}), including 
computational time needed for recolouring, quantitative  assessments and perceptual user studies. We show the usability of our approach for creating visual effects (Section
\ref{sec:usability}) and  conclude (Section \ref{sec:conclusion}).

\section{Related works}
\label{sec:soa}

The body of work in recolouring is very large and the reader is
referred to this recent exhaustive review by Faridul et al. \cite{Lefebvre2014}. 
We focus on presenting landmark methods  (Section \ref{sec:soa:colour:transfer}) as well as the
latest techniques in the field of colour
transfer that  will be used for comparison with our
approach.   Shape registration techniques based on Gaussian
mixture models are then presented in Section
\ref{sec:shape:registration}
as background information for our technique, presented in Section \ref{sec:our:approach}.

\subsection{Colour transfer techniques}
\label{sec:soa:colour:transfer}

Early work on colour transfer started with registering  statistical moments
of colour distributions (Section \ref{sec:soa:registration:moments})
using a parametric affine formulation of the transfer function $\phi$.
This methodology soon shifted to using the optimal transport framework (Section
\ref{sec:soa:optimal:transport}) used in its simplest form for
histogram equalisation and  for correcting
flicker in old (grey scale) films   \cite{NaranjoICIP2000}. 
Optimal transport techniques use non-parametric estimates of
colour distributions, and the resulting algorithms for colour
corrections do not provide an explicit expression of $\phi$ but
instead an estimated correspondence $(x,\phi(x))$ for every colour
pixel $x$. This can be memory consuming when capturing $\phi$ on a $256^3$
discrete RGB colour space for instance.
Alternative  methods (Section \ref{sec:soa:gmm}) instead propose to
capture colour distributions with Gaussian Mixture Models and use
some correspondences (Section \ref{sec:soa:correspondences}) between Gaussian components of the palette and
target distributions.   

\subsubsection{Registration of colour statistical moments}
\label{sec:soa:registration:moments}

The pioneering work of Reinhard et al. \cite{Reinhard2001} proposed to use a warping function $\phi$ with parametric form \cite{Reinhard2001}:
\begin{equation}
\phi(x,\theta=\lbrace \mathrm{G},o\rbrace)= \mathrm{G}\ x+ o
\end{equation}
with the vector $o$ representing an offset  and the $3\times 3$ diagonal matrix $\mathrm{G}$ representing the gains for each colour channel.
The estimation of the parameter $\theta$  is performed by registering the probability density functions of the colour in the  palette and target images, denoted $p_p$ and $p_t$, represented as simple multivariate Gaussians ($p_p\equiv\mathcal{N}(x;\mu_p,\Sigma_p)$ and $p_t\equiv\mathcal{N}(x;\mu_t,\Sigma_t)$) with diagonal   covariance matrices $\Sigma_p$ and $\Sigma_t$. 
Since Normal distributions are fully described by their first two statistical moments, means and covariance matrices, 
the optimal mapping $\phi$ is specified by the solution $\hat{\theta}$ that maps the empirical  estimates of $(\mu_t, \Sigma_t)$ computed using the pixels values $\lbrace x_t^{(i)} \rbrace_{i=1,\cdots,n_t}$ in the target image, to   
the empirical  estimates of $(\mu_p, \Sigma_p)$ computed using the pixels values $\lbrace x_p^{(i)} \rbrace_{i=1,\cdots,n_p}$ in the palette image.

\begin{table*}[!t]
\begin{tabular}{|p{.10\linewidth}|p{0.02\linewidth}|p{.025\linewidth}|p{.025\linewidth}|p{.38\linewidth}|p{.025\linewidth}|p{.025\linewidth}|p{.08\linewidth}|p{.08\linewidth}|}
\hline
Method Name & Ref.& \scriptsize{Corr.} & \scriptsize{No Corr.}& Code availability for testing & \scriptsize{Image} & \scriptsize{Video} &  \scriptsize{Test $P\simeq T$ (Sec. \ref{sec:same:content})}  & \scriptsize{Test $P\neq T$ (Sec. \ref{sec:different:content}) }\\
\hline
\hline
Ours &  & yes & yes  &  \scriptsize{\url{https://www.scss.tcd.ie/~mgrogan/colourtransfer.html}} & yes & yes & yes & yes\\
\hline
\hline
Bonneel  & \cite{Bonneel2015} & no & yes& \scriptsize{\url{https://github.com/gpeyre/2014-JMIV-SlicedTransport}} & yes & no & yes & yes\\
\hline
PMLS  &\cite{Hwang2014} & yes & no & \scriptsize{Results for this method in this paper have been processed by the authors of \cite{Hwang2014} }  & yes & yes &  yes & no\\
\hline 
Ferradans &\cite{Ferradans2013}  & no & yes & \scriptsize{\url{https://github.com/gpeyre/2013-SIIMS-regularized-ot}}  & yes & no  & no & yes\\
\hline
Piti\'{e} &\cite{Pitie2007} & no & yes & \scriptsize{\url{https://github.com/frcs/colour-transfer}} & yes  & no & yes & yes \\
\hline
\end{tabular}
\caption{Road-map for experiments. Our framework is able to take advantage of correspondences (Corr) between Target (T) and Palette (P) images when available, and without correspondences (No Corr). While correspondences are easily available when palette and target images capture the same visual content ($P\simeq T$), they are not available when using images of different content ($P\neq T$). }
\label{tab:roadmap}
\end{table*}

\subsubsection{Optimal transport}
\label{sec:soa:optimal:transport}

Monge's formulation \cite{VillaniBook2009} sets the deterministic decoupling  $y=\phi(x)$ linking two random variables $y \sim p_{t}(y)$ and $x \sim p_{p}(x)$  
imposing the solution $\phi$  that verifies \cite{Villani2003}:
\begin{equation}
p_p(x)=p_t(\phi(x)) \times  |\det\nabla \phi(x)|
\label{eq:solution:gen}
\end{equation}
In practice, finding $\phi$ such that equation (\ref{eq:solution:gen})
is true is difficult  when considering
multidimensional space \cite{Villani2003}. A solution when $x\in \mathbb{R}$ and 
$y\in \mathbb{R}$ can however be defined with  the cumulative distribution of colours in the target and palette images $P_t$ and $P_p$:
\begin{equation}
\phi(x)=P_{t}^{-1}\circ P_{p}(x)
\label{eq:ot:1d}
\end{equation}
when $P_{t}$ is strictly increasing (i.e. $p_t(y) > 0, \forall y\in \mathbb{R}$).
$P_t$ and $P_p$ can be approximated with cumulative histograms  for instance. Such a process for finding the warping function $\phi$ is powerful since no strong hypotheses are made about the distributions (as opposed to  the Gaussian assumption in Section \ref{sec:soa:registration:moments}). 
Moreover, no parametric form is imposed on the warping function $\phi$.

 Solution to Eq. (\ref{eq:solution:gen}) becomes non trivial in multidimensional colour spaces. 
 Of particular interest  is the pioneering work of Piti\'{e} et al. \cite{Pitie2005} who 
proposed an iterative algorithm that first projects the colour pixels  $\lbrace x^{(k)}\rbrace$ on a 1D Euclidean space, and then  estimate $\hat{\phi}$ using  Eq. (\ref{eq:ot:1d}) and apply it to  move all values $\lbrace x^{(k)}\rbrace$ along the direction of the 1D space. 
This operation is repeated with different directions in 1D space until convergence.
The intuitive explanation for this approach comes from the Radon transform  \cite{Pitie2007}: if equation (\ref{eq:ot:1d})  is verified in all 1D projective spaces, then 
Equation (\ref{eq:solution:gen}) will  be verified in the multidimensional space. 
Bonneel et al.  \cite{Bonneel2015} recently proposed to  use a similar radon transform inspired strategy for colour transfer in their optimal transport method.
Their approach is a generalisation of the method proposed by Piti\'{e} et al.  which uses 1-D Wasserstein distances to compute the barycentre of a number of input measures.  As well as being used for colour transfer between target and palette images, their method can also be used to find the barycentre of three or more weighted image palettes. 

Optimal transport is now a widely used framework for solving colour transfer as it allows modelling more various, realistic and complex forms for the distributions $p_t$ and $p_p$  than simple multivariate Gaussians, and allows for a parameter free form of the warping function $\phi$ to be estimated.     
Histograms are  often employed to approximate the colour distributions of images \cite{Neumann2005,Papadakis2011,Pouli2011}
and used in optimal transport methods  \cite{Freedman2010,Ferradans2013,Pouli2011}. Similar to the original Piti\'{e} algorithm  \cite{Pitie2005}, these discrete methods have a tendency to introduce grainy artifacts in the gradient of the result image. 
Piti\'{e} et al's subsequent extension \cite{Pitie2007} proposed a post processing step  to correct this artifact and ensure the gradient field of the recoloured target image is as close as possible to the original target image. 
Similarly, recent methods have  proposed adding a step to impose that the resulting spatial gradient of the recoloured image remains similar to the target image \cite{Xiao2009,Papadakis2011,Ferradans2013}. 
Alternatively other methods have proposed to relax the constraint that enforces the distributions  of the recolored target image and palette image to match exactly \cite{Freedman2010,Pouli2011,Ferradans2013}. 
Bonneel et al.  \cite{Bonneel2015}  also propose using a gradient smoothing technique to reduce any quantisation errors that appear in their results \cite{Rabin2011}.  
Frigo et al. \cite{Frigo15} propose to remove artifacts by first estimating an optimal transport solution and using it to compute a smooth Thin Plate Spline (TPS)  transformation to ensure that a smooth parametric warping function is used for recolouring allowing them to apply their method to video content easily.

\subsubsection{Using Gaussian Mixture models}
\label{sec:soa:gmm}

Jeong and Jaynes \cite{Jeong2007} use colour transfer techniques to
harmonise the colour distributions of non overlapping images of the
same object for tracking purposes in a multiple camera setting.
 Colour chrominance (2D) distribution is modelled using
Gaussian Mixture models, and the transfer function is parametric with
an  affine  form estimated by  minimising the Kullbach-Leibler
divergence between Gaussian Components with a robust
procedure to tackle outlier pairs.
Xiang et al. \cite{Xiang2009682} model the colour distribution of the
target image with a Gaussian mixture estimated by an EM algorithm. 
Each Gaussian component in the mixture defines a local region in the
target image, and each segmented local region is recoloured independently using
multiple palette-source images, likewise segmented in regions using GMM
fitting of their colour distributions. Colour transfer is performed by
associating the best Gaussian component from the sources to the 
Gaussian target region using Reinhard's Gaussian affine transfer function
\cite{Reinhard2001}.  
Segmentation using alternative approaches (i.e. Mean Shift \cite{Comaniciu2002}, K-means \cite{XuICIP2005}) is
also tested to define the colour Gaussian mixtures. This approach
relies on homogeneous colour regions, each  captured with one
multivariate normal in the 3D colour space.
Localised colour transfer using Gaussian Mixture models between
overlapping colour images have also been proposed for colour corrections,
motion deblurring, denoising, gray scale coloring \cite{TaiCVPR2005}.

\subsubsection{Using correspondences}
\label{sec:soa:correspondences}

Using one to one correspondence between a Gaussian component capturing the colour content of the  target image  and palette images (or their regions) have been used for estimating an affine warping function $\phi$. Other methods which take correspondences into account when finding the colour transformation include \cite{Oliveira2015,HaCohen2011,Hwang2014}.
 Oliveira et al. \cite{Oliveira2015} proposed to find the mapping of 1-dimensional truncated GMM representations computed for each colour channel of the target and palette images. Hwang et al. \cite{Hwang2014} use moving least squares to estimate an affine transformation for each pixel in the target image with the least squares algorithm. To tackle sensitivity to outlier correspondences, only  subsets of the correspondences are used as control points and  a probabilistic framework 
is deployed to remove correspondences that are likely to be incorrect. 
However both of these methods are only applicable when the correspondences between target pixels and palette pixels are available.

\subsection{Shape registration}
\label{sec:shape:registration}

\textcolor{black}{In this section, we consider shapes as entities mathematically described as a set
of $n$ contour points in 2D, and as a set of $n$ vertices on the surfaces of   objects in 3D \cite{Jian2011}.
A shape is hence  denoted as a point cloud $\lbrace x^{(i)} \rbrace_{i=1,\cdots,n}$, where the point description  $x^{(i)}$ corresponds  to the 2D or 3D coordinates in the spatial domain.  This description can, however,  be augmented with additional information such as the local normal direction to the contour or surface \cite{ArellanoPR2015}.}
In the past decade or so, several shape registration techniques have
been formulated as minimising a divergence between probability density
functions capturing the target shape,  $p_t(x)$, and  a
 model shape, $p_p(x)$, by finding the parametric  deformation
$\phi$ to apply to the target shape to register it on the model shape (Section \ref{sec:soa:phi:tps}).
The probability density functions $p_t$ and $p_p$ can be approximated by kernel density estimates using
observations
$\lbrace x_t^{(i)} \rbrace_{i=1,\cdots,n_t}$ and $\lbrace x_p^{(i)} \rbrace_{i=1,\cdots,n_p}$ describing the two shapes to be registered. Several kernels, such as the Gaussian kernel, are easier to use for registration (Section \ref{sec:soa:shape:gmm}).
Many dissimilarity metrics  have been defined for pdfs and the  Kullbach-Leibler divergence, already   mentioned to solve  colour transfer,  is probably the most well known  \cite{ITBook2009}.  Section \ref{sec:divergence} proposes instead a more robust alternative to Kullback Leibler divergence for comparing pdfs that corresponds to the Euclidean $\mathcal{L}_2$ distance  between these pdfs. 
Some correspondences between the target and model shapes, denoted $\lbrace (x_t^{(k)},x_p^{(k)})\rbrace_{k=1,\cdots,n}$ with $n\leq n_t$ and $n\leq n_p$, may occasionally be available and can also be efficiently used  with the $\mathcal{L}_2$ distance (Section \ref{sec:soa:shape:correspondences}).

\subsubsection{$\mathcal{L}_2$ distance}
\label{sec:divergence}

The Euclidean $\mathcal{L}_2$ distance between two pdfs is defined as:
\begin{equation}
\mathcal{L}_2=\| p_t-p_p\|^2 = \int (p_t(x)-p_p(x))^2 dx
\end{equation}
and it can be conveniently rewritten $\mathcal{L}_2=\| p_t\|^2-2 \langle p_t|p_p\rangle + \|p_p\|^2 $.
Of note the term $\| p\|^2$ is proportional  to the quadratic Renyi entropy of $p$ (for target and palette),
while the scalar product term $\langle p_t|p_p\rangle $ is proportional  to the Renyi cross entropy between $p_t$ and $p_p$ \cite{ITBook2009}.
The advantage of computing $\mathcal{L}_2$ over the Kullbach Leiber divergence between probability density functions is that it can be computed  explicitly with Gaussian mixtures and it has  also been shown to be more robust to outliers \cite{Jian2011,ScottTechnometrics2001}.

\subsubsection{GMM representation of shapes}
\label{sec:soa:shape:gmm}

 Gaussian Mixture Models have been a popular representation of shapes  for solving  shape registration  
\cite{ITBook2009}. 
Recently  Jian et al. \cite{JianICCV2005,Jian2011} proposes to register two point cloud shapes  $\lbrace x_t^{(i)} \rbrace_{i=1,\cdots,n_t}$ and $\lbrace x_p^{(i)} \rbrace_{i=1,\cdots,n_p}$, by first fitting a kernel density estimate to each point set using the Gaussian kernel. 
The target distribution is formulated as \cite{JianICCV2005,Jian2011}:
\begin{equation}
p_t(x|\theta)=\frac{1}{n_t} \sum_{i=1}^{n_t} \mathcal{N}(x;\phi(x_t^{(i)},\theta),h^2\mathrm{I}) 
\end{equation}
with notation $\mathcal{N}(x;\mu,\Sigma)$ indicating a Normal distribution for random vector $x$, with mean $\mu$ and covariance $\Sigma$.
The function $\phi(x,\theta)=\phi_{\theta}(x)$ moves vertex $x_t^{(i)}$ and
this displacement is controlled by a latent vector $\theta$ to be
estimated. 
The second kernel density estimate is defined using the second point set as:
\begin{equation}
p_p(x)=\frac{1}{n_p} \sum_{i=1}^{n_p} \mathcal{N}(x;x_p^{(i)}, h^2 \mathrm{I}). 
\end{equation}
Isotropic covariance matrices controlled by a bandwidth $h$ (with $\mathrm{I}$ a $3\times 3$ identity matrix) are used for both distributions.
The estimation of the warping function $\phi$ consists of finding $\theta$ that minimises the $\mathcal{L}_2$ distance between $p_p$ and $p_t$:
\begin{equation}
\hat{\theta}=\arg\min_{\theta} \left \lbrace \| p_t\|^2-2 \langle p_t|p_p\rangle \right\rbrace
\end{equation}
since $\|p_p\|^2$ does not depend on $\theta$ \cite{ScottTechnometrics2001}. Terms $\| p_t\|^2$ and $\langle p_t|p_p\rangle$ can be computed explicitly using the integral result for Normals: 
$$
\int \mathcal{N}(x;\mu_1,\Sigma_1 ) \ \mathcal{N}(x;\mu_2,\Sigma_2)\ dx = \mathcal{N} (0;\mu_1-\mu_2, \Sigma_1+\Sigma_2)
$$

\subsubsection{Deformable transfer function $\phi$}
\label{sec:soa:phi:tps}

The transfer function $\phi: x\in \mathbb{R}^d \rightarrow
\phi_{\theta}(x)\in \mathbb{R}^d$  can be conveniently defined as a rigid, affine or
Thin Plate Splines parametric function \cite{ITBook2009,Jian2011} which can be written as:
 \begin{equation}
    \phi_{\theta}(x) = \underbrace{\mathrm{A}\ x + o}_{\text{Affine}}
    + \underbrace{\sum^m_{j = 1} w_j \ \psi(\|x - c_j \|
      )}_{\text{nonlinear}} = \mathrm{A}\ x + o +\mathrm{W} \ \Psi(x)
\label{eq:orig_tps}
\end{equation}
Discarding the non linear part of $\phi$ provides an affine warping function which can be further limited to a rigid tranformation by choosing $\mathrm{A}$ as a rotation matrix. 
For the more general TPS form, the latent parameters of interest, $\theta$, corresponds to the set of
coefficients $\lbrace w_j\in \mathbb{R}^d\rbrace$,  the $d\times d$ matrix
$\mathrm{A}$ and the $d$-dimensional
vector offset $o$. The non linear part can be rewritten in matrix form
with a $d\times m$ matrix $\mathrm{W}$ gathering coefficients $\lbrace
w_j\in \mathbb{R}^d\rbrace$, and an $m$ dimensional vector $\Psi(x)$
collating all values $\lbrace \psi(\|x - c_j \|) =-\|x - c_j \|\rbrace$.
The set of control points $\{c_j \}_{j=1,..,m}$ are a subset of the
observations $\lbrace x^{(i)}\rbrace$ \cite{Jian2011}.
Other deformable models (linear in $x$ and $\theta$)  have been used with $\mathcal{L}_2$ to register shapes \cite{Eusipco2012Arellano1,CVMP2013Arellano}.
These models rely on Principal Component Analysis to learn  deformations of a family of shapes. 
Regularisation terms can also  be added to $\mathcal{L}_2$ to constraint the estimation of $\theta$  \cite{Jian2011,Eusipco2012Arellano1}.

\subsubsection{Using correspondences}
\label{sec:soa:shape:correspondences}

Finding  correspondences between shapes can be done using  local descriptors such as
shape context for point clouds \cite{BelongiePAMI2002}, or using  SIFT
features \cite{LoweIJCV2004} when
considering images for instance. 
Ma et al.  \cite{MaCVPR2013} have proposed to use only the correspondences for performing shape registration with $\mathcal{L}_2$ to improve both robustness and computational efficiency. 


\subsection{Remarks}

\textcolor{black}{Shapes described as point clouds in the spatial domain (2D or 3D) have the same mathematical representation 
as colour  pixels  values of an image represented as points in a 3D colour space. 
Therefore the methods presented in Section \ref{sec:shape:registration} for registering point cloud shapes 
can be used to register point clouds of colour triplets of a target and palette images.  
This is the core idea of this paper with more details for its implementation given in paragraph \ref{sec:our:approach}.}

\textcolor{black}{Piti\'{e} et al. have also shown experimentally that their original optimal transport algorithm for colour transfer iteratively decreases the Kullbach Leibler divergence between the probability density functions $p_t$ and $p_p$ \cite{Pitie2007}. 
As an alternative, we  instead propose to explicitely formulate the problem of colour transfer as minimising a divergence ($\mathcal{L}_2$) between pdfs.}

\textcolor{black}{Piti\'{e} et al's non parametric formulation of the solution $\hat{\phi}$ gives, for each pixel $x_t^{(i)}$, a corresponding value $\hat{\phi}(x_t^{(i)})$ but does not provide  the possibility of applying this transformation to a previously unseen value $x$ \cite{Pitie2007}. Moreover  the warping function $\hat{\phi}$ cannot be  directly manipulated for mixing with other estimated warping functions. 
Alternatively, a parametric affine warping function $\phi_{\theta}$ can be  computed directly  with the optimal transport framework  \cite{PitieRigidColourTransfer2007}.
However this computation assumes that  target and palette pdfs are Normals which is a gross approximation of colour distributions in images in general. Moreover affine warping functions are limited for registering colour statistics, and  as an improvement, Frigo et al. \cite{Frigo15} has proposed instead to fit a smooth Thin Plate Spline (TPS)  transformation to the non parametric optimal transport solution to propose a more complex non rigid warping function $\phi_{\theta}$ suitable for colour transfer.}

\section{Robust Colour Transfer}
\label{sec:our:approach}

\subsection{GMM representation of colour content}

We define two Gaussian mixture models  $p_t(x)$ and $p_p(x)$ capturing  the colour content of the target and palette image respectively as follows (omitting subscript $(t,p)$): 
\begin{equation}
p(x)= \sum_{k=1}^{K}  \ \mathcal{N}(x;\mu^{(k)},\Sigma^{(k)}) \ \pi^{(k)} 
\label{eq:gmm}
\end{equation}
with $K\in\lbrace K_t,K_p\rbrace$ the number of Gaussians in the mixture for the  target and palette images, $\mu^{(k)}\in\lbrace \mu^{(k)}_t,\mu^{(k)}_p\rbrace$ the respective means and $\Sigma^{(k)}\in\lbrace \Sigma^{(k)}_t,\Sigma^{(k)}_p\rbrace$
  covariances of the  Gaussian $k$.  The coefficients $\pi^{(k)} \in\lbrace \pi_t^{(k)} ,\pi_p^{(k)} \rbrace $ are positive weights, all summing to one, capturing the relative importance of each Gaussian component in the mixture. 
 The target distribution $p_t(x)$ is changed to a parametric family of distributions $p_t(x|\theta)$ by changing the means $\lbrace \mu^{(k)}_t\rbrace$ to $\lbrace \phi(\mu^{(k)}_t,\theta) \rbrace$. The  Renyi cross entropy term in $\mathcal{L}_2$ is then proportional to: 
\begin{equation}
\langle p_t|p_p \rangle=
\sum_{k=1}^{K_t} \sum_{l=1}^{K_p} 
\mathcal{N}(0;\phi(\mu_t^{(k)}, \theta)-\mu_p^{(l)},\Sigma_t^{(k)}+\Sigma_p^{(l)}) \ \pi_t^{(k)}  \ \pi_p^{(l)} 
  \label{eq:correspondencesno}
\end{equation}
while the quadratic entropy of $p_t$ is proportional to: 
\begin{equation}
\| p_t \|^2=
\sum_{k=1}^{K_t} \sum_{l=1}^{K_t} 
\mathcal{N}(0;\phi(\mu_t^{(k)}, \theta)-\phi(\mu_t^{(l)}, \theta),\Sigma_t^{(k)}+\Sigma_t^{(l)}) \ \pi_t^{(k)}  \ \pi_t^{(l)} 
  \label{eq:renyi:entropy:gmm}
\end{equation}
Jian et al. use $\mathcal{L}_2$ to register shapes encoded as Gaussian Kernel Density Estimates \cite{Jian2011}. Using this idea for our colour transfer application would mean that the number  $K$ of Gaussians in the mixture (Eq. \ref{eq:gmm}) is the number of pixels in the image (target or palette), and the cluster centres $\mu^{(k)}=x^{(k)}$ are   the colour pixels $x^{(k)}
$ in the image. In this case $K$ is extremely large and 
our cost function becomes too computationally intensive to be practical. 
As an alternative two algorithms for clustering, K-means and Mean Shift,  are proposed to compute the means 
(Sections \ref{sec:kmeans} and \ref{sec:ms}). When correspondences are available between palette and target images, a  third  method for setting the means is proposed in Section 
\ref{sec:correspondences}.

\subsubsection{Covariances}
Both target and palette GMMs uses isotropic identical covariance matrices controlled by a bandwidth $h$:
\begin{equation}
\Sigma=h^2 \mathrm{I}
\label{eq:bandwidth:covariance}
\end{equation}
with the identity matrix $\mathrm{I}$. In this case the scalar product between two Normal distributions is:
\begin{equation}
\langle \mathcal{N}(\mu_1,h^2 \mathrm{I}) | \mathcal{N}(\mu_2,h^2 \mathrm{I})\rangle =\mathcal{N}(0;\mu_1-\mu_2,2h^2 \mathrm{I})
\end{equation}
and when $h$ is very large
the Taylor series expansion of $e^{-x}$ can be used to compute the approximation: 
\begin{equation}
e^{\frac{-(\mu_1 - \mu_2)^2}{4h^2}} \simeq 1-\frac{1}{4h^2}(\mu_1-\mu_2)^2
\end{equation}

Finding $\theta$ corresponds in this case to minimising the quadratic convex function:
\small
\begin{multline}
\|p_t\|^2-2\langle p_t|p_p \rangle\simeq -\sum_{k=1}^{K_t} \sum_{l=1}^{K_t}  \left( \frac{\phi(\mu_t^{(k)}, \theta)-\phi(\mu_t^{(l)}, \theta)}{2h} \right)^2 \ \pi_t^{(k)}  \ \pi_t^{(l)} \\
 +2 \sum_{k=1}^{K_t} \sum_{l=1}^{K_p}  \left( \frac{\phi(\mu_t^{(k)}, \theta)-\mu_p^{(l)}}{2h} \right)^2 \ \pi_t^{(k)}  \ \pi_p^{(l)} -1 
\end{multline}

\normalsize
This cost function is a trade off between moving  $\mu_t$s as close as possible to $\mu_p$s (Renyi cross entropy term), while avoiding collapsing all $\mu_t$s to a sole mean vector (quadratic Renyi entropy term). In other words, the target pdf $p_t$ is prevented from collapsing into a unique Gaussian distribution and for keeping the $p_t$ as  spread as possible over the colour space.
The bandwidth $h$ is also used as a temperature in our algorithm (starting with a large value of $h$) when estimating  $\theta$ in a simulated annealing framework to avoid local solutions.

\subsubsection{Weights}
The natural choice for setting the weights $\pi^{(k)}$ is to choose the proportion of pixels in the image associated with the cluster $k$.
However since target and palette images do not have exactly the same visual content in general, as in \cite{Frigo15} we found that matching the colour content of the images was preferable to matching the true colour distribution of the images. Therefore we choose equal weights for each cluster and  set $\pi^{(k)}=\frac{1}{K}$.

\subsection{K-means Algorithm}
\label{sec:kmeans}
The K-means clustering algorithm can be used to define
$\lbrace\mu^{(k)}\rbrace$ for both target and  palette images
\cite{Grogan2015,Grogan2015b}. The computational complexity is then
controlled when choosing the number $K$ of clusters in the pixels
values sets $\lbrace x_t^{(i)}\rbrace_{i=1,\cdots,n_t}$ and $\lbrace
x_p^{(i)}\rbrace_{i=1,\cdots,n_p}$. The K-means algorithm  is equivalent to using an EM algorithm enforcing identical isotropic covariance matrices \cite{Wu2007}.

\subsection{Mean Shift Segmentation}\label{sec:ms}
\textcolor{black}{A possible  alternative to K-means is the  Mean Shift clustering algorithm which has been previously proposed in other colour transfer techniques for image segmentation \cite{Oliveira2015,Xiang2009682} and takes into account spatial information when computing the cluster centres   $\lbrace\mu^{(k)}\rbrace$ \cite{Comaniciu2002}\footnote{Code for \cite{Comaniciu2002}   at \url{http://coewww.rutgers.edu/riul/research/code.html}}.} The Mean Shift algorithm can also be thought of as an EM algorithm \cite{Carreira-PAMI07}.  


\subsection{Correspondences}
\label{sec:correspondences}

 Pixel pairs between target and palette images, denoted  $\{(x_t^{(k)}, x_p^{(k)} ) \}_{k = 1,\cdots,n}$, can be   computed  when considering target and palette images capturing the same scene. Colour transfer techniques are indeed often used in this context to harmonise colours across a video sequence and/or across multiple view images such as in the bullet time visual effect.
In this case, the means of the Gaussian mixtures are set such that
$\{(\mu_t^{(k)}, \mu_p^{(k)} )=(x_t^{(k)}, x_p^{(k)} ) \}_{k = 1,\cdots,n}$ imposing $K_t=K_p=n$ when defining the distributions $p_t$ and $p_p$. 
Moreover, the scalar product $\langle p_t|p_p \rangle$ in  our cost function is then simplified as follows: 
\small
\begin{equation}
\begin{split}
\langle p_t|p_p \rangle= \sum_{k=1}^{n} \mathcal{N}(0;\phi(\mu_t^{(k)}, \theta)-\mu_p^{(k)},\Sigma_t^{(k)}+\Sigma_p^{(k)})  \  \pi_t^{(k)} \ \pi_p^{(k)}
  \label{eq:correspondences}
  \end{split}
\end{equation}
\normalsize
The computational complexity of this term is then $n=K_t=K_p$ when using $n$ correspondences, and $K_t\times K_p$ without correspondences.
Performance of our approach with correspondences is assessed  in Section \ref{sec:quant:results} using palette and target images with similar content. Pixel correspondences are not used when considering palette and target images with different content but K-means or Mean Shift clustering are used instead.

\subsection{Warping function $\phi$}
\label{sec:warping}

\begin{table}[!h]
\begin{center}
 \begin{tabular}{|l|l|}
 \hline
   \textbf{RBF name}  &\textbf{RBF equation} \\ \hline
    TPS & $\psi(r)=-(r)$\\ 
  
   Gaussian & $\psi(r)=e^{-(\epsilon r)^2}$  \\ 
   Inverse Multiquadric & $\psi(r)=\frac{1}{\sqrt{1+(\epsilon r)^2}}$  \\ 
   Inverse Quadric & $\psi(r)= \frac{1}{{1+(\epsilon r)^2}}$ \\ \hline
 \end{tabular}
 \end{center}
    \caption{The equations for the different radial basis functions $\psi$ tested in this paper. }
    \label{tab:RBF_eq}
\end{table}

In this paper, several radial basis functions $\psi$ are tested including TPS to define the warping function $\phi$ (cf. Eq. \ref{eq:orig_tps}).
Possible choices for $\psi$ are  listed in Table \ref{tab:RBF_eq}.
The colour spaces considered for testing our framework are the RGB and
Lab spaces. In both cases, the $m$ control points  $\lbrace c_j
\rbrace_{j=1,\cdots,m}$ are chosen on a regular grid spanning the 3D
colour space such that the 3D grid has $m=5\times
5\times 5=125$ control points in the colour space. As a consequence
the dimension of the latent space
to explore for the estimation of $\theta$
 is:
$$
\dim(\theta) = 125 \times 3+9+3=387
$$ 
with $\dim(c_j)=3, \ \forall j$, $\dim(A)=3\times 3=9$ and $\dim(o)=3$ (cf. Eq. \ref{eq:orig_tps}).

\subsection{Estimation of $\theta$}
\label{sec:our:theta:estimation}

To enforce that a smooth solution $\phi$ is estimated, the Euclidean distance
$\mathcal{L}_2(\theta)$ is minimised with a roughness penalty on the transfer function $\phi$ \cite{ITBook2009} and the estimation is performed as:
\begin{equation}
\hat{\theta}=\arg\min_{\theta} \left \lbrace\mathcal{C}(\theta)= \|p_t\|^2-2\langle p_t|p_p \rangle + \lambda  \int  \| D^2  \phi(x,\theta)\|^2 dx \right \rbrace
\label{eq:estimation:theta}
\end{equation}
with $\| D^2  \phi(x,\theta)\|^2=\sum_{i,j \in \lbrace 1,\cdots, d\rbrace }\left(\frac{\partial^2 \phi}{\partial x_i \partial x_j}\right)^2$ with $d=3$ for the colour spaces.

\subsection{Implementation Details}
\label{sec:implementation}

Our strategies to estimate $\theta$ are summarised in Algorithm \ref{algo:estimation}.
\begin{algorithm}[H]
\begin{algorithmic}
\Require $\hat{\theta}$ initialised so that $\phi_{\hat{\theta}}(x)=x$ (identity function)
\Require  $hmin$, $hmax$, $\lambda$ and choose $\psi$ (with  $\epsilon$)
\Require Initialisation of $\lbrace \mu_t^{(i)}\rbrace$, $\lbrace \mu_p^{(i)}\rbrace$ or $\lbrace (\mu_t^{(i)}, \mu_p^{(i)})\rbrace$ (correspondences)
\If{Using K-means (no correspondences)}
\State choose $K_t$ and $K_p$
\State Cost function $\mathcal{C}(\theta)$ (Eq. \ref{eq:estimation:theta}) defined with  term $\langle p_t|p_p\rangle$ (Eq. \ref{eq:correspondencesno})
\ElsIf{Using Mean Shift (no correspondences)}
\State $K_t$ and $K_p$ are estimated by Mean Shift
\State Cost function $\mathcal{C}(\theta)$ (Eq. \ref{eq:estimation:theta})   defined with term  $\langle p_t|p_p\rangle$ (Eq. \ref{eq:correspondencesno})
\ElsIf{Using $n$ correspondences}
\State $K_t=K_p=n$
\State  Cost function $\mathcal{C}(\theta)$ (Eq. \ref{eq:estimation:theta}) defined with term  $\langle p_t|p_p\rangle$ (Eq. \ref{eq:correspondences})
\EndIf
\State Start $h=hmax$ (controls covariance matrix Eq. \ref{eq:bandwidth:covariance})
\Repeat 
\State $ \hat{\theta} \leftarrow\arg\min_{\theta} \mathcal{C}(\theta)$ 
\State $h\leftarrow .5 \times h$ (annealing)
\Until{Convergence $h<hmin$}
\Return $\hat{\theta}$ 
\end{algorithmic}
\caption{Our strategies for estimating the parameter $\theta$ for the warping function $\phi_{\theta}(x)$.}
\label{algo:estimation}
\end{algorithm}
Two values, $\lambda$ and $\epsilon$ (in the case of the Gaussian, Inverse Quadric and Inverse Multiquadric basis functions as defined in Tab. \ref{tab:RBF_eq}) need to be chosen in our framework.  
Table \ref{tab:paramChosen} gives the values that were used in our experiments to get the best results overall.
 \begin{table}[!h]
\begin{center}
 \begin{tabular}{|c|c|c|c|c|}
 \hline
  RBF $\psi$ and colour space &  \multicolumn{2}{|c|}{Correspondences} & \multicolumn{2}{|c|}{No Correspondences } \\ \hline
  & $\lambda$ & $\epsilon $ & $\lambda$ & $\epsilon $ \\ \hline
 TPS$_{rgb}$ & $3e^{-3}$ & \ding{55} &  $3e^{-6}$  &   \ding{55}  \\ \hline
 
 TPS$_{lab}$ & $3e^{-3}$ &  \ding{55} & $3e^{-4}$ &   \ding{55} \\ \hline
 
 G$_{rgb}$ & $3e^{-5}$ & $6e^{-3}$ & $3e^{-8}$ & $6e^{-3}$\\ \hline
 
 G$_{lab}$  & $6e^{-3}$ & 3 & $3e^{-4}$ & 3 \\ \hline
  
InMQ$_{rgb}$ & $3e^{-5}$ & $6e^{-3}$& $3e^{-8}$ & $6e^{-3}$\\ \hline

InMQ$_{lab}$ & $6e^{-3}$ & 10 & $3e^{-4}$ & 3  \\ \hline

InQ$_{rgb}$ &$3e^{-6}$ & $6e^{-3}$ &  $3e^{-8}$ & $6e^{-3}$  \\ \hline
 
InQ$_{lab}$ & $6e^{-3}$ & 30 & $3e^{-4}$  & 3   \\ \hline
 \end{tabular}
 \end{center}
    \caption{Selected parameters $\lambda$ and $\epsilon$ in our algorithms tested in our experiments in Section \ref{sec:quant:results}. In addition, the number of clusters for K-means  is set to $K_t=K_p=50$ , while the number of correspondences $n=50000$  are chosen randomly across the overlapping region of the target and palette images after registration. When Mean Shift clustering is used, the number of clusters ($K_t$ and $K_p$) are estimated along with the cluster centres.}
    \label{tab:paramChosen}
\end{table}
For clarity we extend that notation in Table \ref{tab:paramChosen} so that  TPS$_{rgb}^{KM}$ indicates the basis function $\psi$  is TPS, the colour space is RGB, and the clustering techniques for finding the means of the GMMs is K-means
(similarly denote TPS$_{rgb}^{MS}$ for Mean Shift and TPS$_{rgb}^{Corr}$ when using Correspondences).

\subsection{Parallel Recolouring step}
\label{sec:parallel}

One of the main advantages of this method is the fact that our transformation is controlled by a parameter $\theta$ and any pixel $x$ can be recoloured by computing the new colour value $\phi_{\theta}(x)$.
This computation can be done in parallel by distributing the pixels to be recoloured to the multiple processors that are available.
Moreover, with one target image  and $N$ palette images of choice, the transformations $\{ \hat{\theta}_1,\cdots, \hat{\theta}_N \}$ can also be estimated in parallel, and any interpolated new value $\theta$
can be used for recolouring:
\begin{equation}
\theta_{new} = \sum_{j=1}^N \gamma_j \ \hat{\theta}_j, \quad \text{with } \sum_{j=1}^N \gamma_j =1,\text{and} \  \gamma_j\geq 0 \  \forall j 
\label{eq:interpolation:theta}
\end{equation}
Interpolating in the $\theta$ space to create a new warping function is made easy thanks to the fact that we chose
control points on a regular grid (cf. Section \ref{sec:warping}) and not as sub samples of palette images (cf. Section \ref{sec:soa:phi:tps}).
Of particular interest are the creation of smooth temporal transitions between the identity warping function and a  colouring warping function for instance.
In Section \ref{sec:visual:effects} we show more on how to create visual effects using interpolation masks.
 Having a quick recolouring step is  essential to give the user instant feedback about the new effects being applied to the image or video. 
Our transformation can be applied to each pixel independently and it is therefore highly parallelisable. A CPU or GPU parallel implementation would ensure that the target image is recoloured almost instantly. 
 For our implementation we parallelised the recolouring step on the CPU using OpenMP, and performance is assessed in  Section \ref{sec:computation:time}.

\section{Experimental results}
\label{sec:quant:results}

To quantitatively assess recolouring results,  three metrics, peak signal to noise ratio (PSNR), structural similarity index (SSIM) and \textcolor{black}{colour image difference (CID)}, 
are often  used when considering palette and target images of the same content for which correspondences are easily available  \cite{Oliveira2015,Hwang2014,Lissner2013}. 
Alternatively user studies have also been used to assess the perceptual visual quality of the recolouring \cite{Hristova2015}.

We evaluate our proposed algorithms and show that they are comparable to current state of the art colour transfer algorithms in terms of the perceptual quality of the results (paragraphs \ref{sec:same:content}, \ref{sec:different:content} and \ref{sec:qualitative:user:study}), and superior in terms of computational speed (paragraph \ref{sec:computation:time}). Moreover our parametric formulation for colour transfer allows easy and flexible manipulations by artists for creating new visual effects (paragraph \ref{sec:visual:effects}). \textcolor{black}{When carrying out our colour transfer algorithm in Lab space \footnote{\textcolor{black}{Matlab functions rgb2lab and lab2rgb with reference white point `d65' used when converting RGB and Lab, and vice versa.}}, both the clustering and registration is performed in Lab space, with the results converted back to RGB before the metrics are computed.}

%

Table \ref{tab:roadmap} summarises  the  methods (including ours) used for comparison in this paper.

\subsection{Images with similar content  $P\simeq T$ }
\label{sec:same:content}

One important application of colour transfer is in harmonising the colour palette of several images or videos capturing the same scene.
To evaluate our algorithm applied to images with similar content, we use the 15 images in the dataset provided by Hwang et al. \cite{Hwang2014}\footnote{\url{https://sites.google.com/site/unimono/pmls}} which includes images with many different types of colour changes including different illuminations, different camera settings and different colour touch up styles. This dataset  provides palette images which have been aligned to match the target image (c.f. Figure \ref{fig:sameContent}). To define correspondences,  pixels at the same location in the target and aligned palette images are selected together to form a pair \cite{Hwang2014}.

\subsubsection{Choice of radial basis function $\psi$}
Table \ref{tab:MyMethods_SSIM_PSNR}  provides quantitative metrics to evaluate our algorithms with correspondences using  the PSNR, SSIM and \textcolor{black}{CID} metrics\footnote{See supplementary material for image results. }.
\begin{table}[!h]
\begin{center}
 \begin{tabular}{|l|l|l|l|l|l|l|}
 \hline
   \multicolumn{1}{|c|}{}& \multicolumn{2}{|c|}{PSNR} &\multicolumn{2}{|c|}{SSIM} &\multicolumn{2}{|c|}{\textcolor{black}{CID}} \\ \hline
    & $\mu$ & $SE$ & $\mu$ & $SE$ & $\mu$ & $SE$ \\ \hline
    TPS$_{rgb}^{Corr}$ & 30.30 & 1.5 & \textcolor{blue}{0.944} &  0.02 & \textcolor{OliveGreen}{0.172} & 0.02\\ \hline
   TPS$_{lab}^{Corr}$  & \textcolor{red}{30.56} & 1.4  & 0.942 & 0.02 & 0.182 & 0.03 \\ \hline
   G$_{rgb}^{Corr}$  & 30.03 & 1.5 &  0.944 & 0.02 & 0.176 & 0.02  \\ \hline
   G$_{lab}^{Corr}$   &  \textcolor{OliveGreen}{30.56} & 1.4 & 0.942 & 0.02 & 0.184 & 0.03 \\ \hline
   InMQ$_{rgb}^{Corr}$  & 30.37  & 1.5 &  \textcolor{red}{0.944} & 0.02 & \textcolor{blue}{0.173} & 0.02\\ \hline
   InMQ$_{lab}^{Corr}$ & \textcolor{blue}{30.49} & 1.4 & 0.942 & 0.02 & 0.183 & 0.03\\ \hline
   InQ$_{rgb}^{Corr}$  & 30.37 & 1.5 & \textcolor{OliveGreen}{0.944} & 0.02 & \textcolor{red}{0.169} & 0.02 \\ \hline
   InQ$_{lab}^{Corr}$ & 30.22  & 1.4 & 0.940 & 0.02 & 0.188 & 0.02\\ \hline
  \end{tabular}
 \end{center}
    \caption{Assessment of our algorithms using correspondences with several basis functions $\psi$ in two colour spaces. The mean PSNR, SSIM and \textcolor{black}{CID} values $\mu$, and standard error $SE$, for each method computed on the 15 images in the dataset. Highest PSNR and SSIM values, and lowest \textcolor{black}{CID} values, indicate the best results(best in red, second best in green, third best in blue).}
    \label{tab:MyMethods_SSIM_PSNR}
\end{table}
In general the methods applied in the Lab colour space provide slightly better results in terms of PSNR, and slightly worse results in terms of SSIM and \textcolor{black}{CID}. A t-test comparing the mean values of each method confirms though that  the difference between them is statistically insignificant (with 99\% confidence). TPS has  the advantage of being faster to compute (paragraph \ref{sec:computation:time}) and since it achieves perceptually similar performance, it is mainly used for comparison in the rest of the paper.

\subsubsection{With or without correspondences}

Table \ref{tab:new} compares TPS$_{rgb}$ with correspondences  (TPS$_{rgb}^{Corr}$) to stress that using correspondences allows the algorithm to outperform our alternative algorithms that do not use them (TPS$_{rgb}^{KM}$ and TPS$_{rgb}^{MS}$). Note however that there is no statistical difference in terms of PSNR, SSIM and \textcolor{black}{CID} between the results of TPS$_{rgb}^{KM}$ using K-means and  TPS$_{rgb}^{MS}$ using Mean Shift on this test where palette and target images capture the same scene. Section \ref{sec:different:content} will show how TPS$_{rgb}^{KM}$ gives perceptually more pleasing results than TPS$_{rgb}^{MS}$ for images of different content.
\begin{table}[!h]
    \centering
    \begin{tabular}{l|lll}
       & TPS$_{rgb}^{Corr}$ & TPS$_{rgb}^{KM}$& TPS$_{rgb}^{MS}$\\
        \hline
        PSNR & \textcolor{red}{30.30} (1.5)& 24.20 (0.8) & 24.47 (0.8)\\
        SSIM & \textcolor{red}{0.944} (0.02)& 0.908 (0.02) & 0.896 (0.02)\\
        \textcolor{black}{CID} & \textcolor{red}{0.172} (0.02)& 0.325 (0.02) & 0.337 (0.03)\\
        \hline
    \end{tabular}
    \caption{The mean PSNR, SSIM and \textcolor{black}{CID} values for the set of images with similar content, with the standard error shown in brackets. Using  correspondences leads to better results.}
    \label{tab:new}
\end{table}

\subsubsection{Robustness of $\mathcal{L}_2$ to outlier correspondences}
\begin{figure*}[!t]
\begin{center}
\begin{tabular}{cccccc}
  &  &  &   &   &  \\
\includegraphics[width = .15\linewidth, height = .09\linewidth]{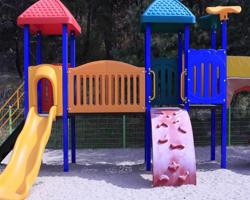} &
\includegraphics[width = .15\linewidth, height = .09\linewidth]{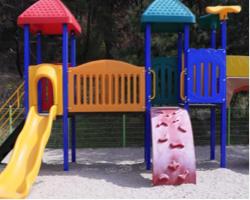} &

\includegraphics[width = .15\linewidth, height = .09\linewidth]{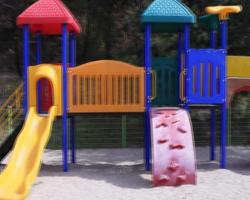} &

\includegraphics[width = .15\linewidth, height = .09\linewidth]{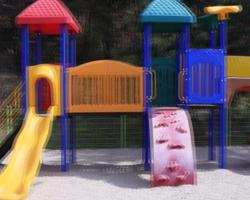} &
\includegraphics[width = .15\linewidth, height = .09\linewidth]{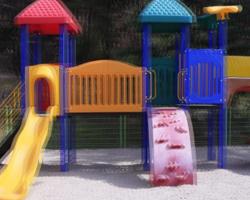}&
\includegraphics[width = .15\linewidth, height = .09\linewidth]{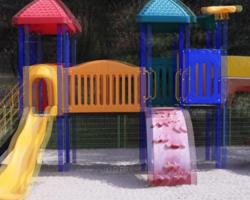} \\
Target $T$ & \small{ $(T-P)_2$} &\small{ $(T-P)_5$} & \small{$(T-P)_8$} & \small{$(T-P)_{10}$}  & \small{$(T-P)_{15}$}  \\ 

\includegraphics[width = .15\linewidth, height = .09\linewidth]{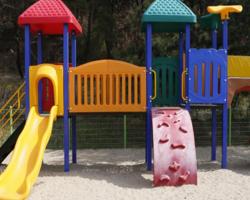} &
\includegraphics[width = .15\linewidth, height = .09\linewidth]{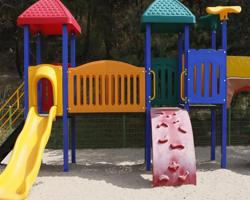} &

\includegraphics[width = .15\linewidth, height = .09\linewidth]{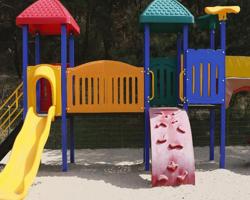} &

\includegraphics[width = .15\linewidth, height = .09\linewidth]{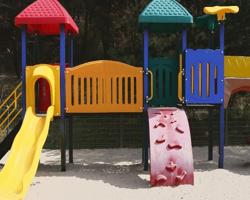} &
\includegraphics[width = .15\linewidth, height = .09\linewidth]{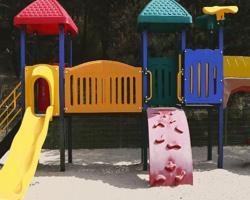}&
\includegraphics[width = .15\linewidth, height = .09\linewidth]{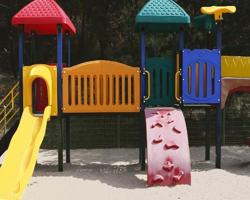} \\
 Palette $P$ & \scriptsize{SSIM $= 0.918$}   &  \scriptsize{SSIM$ = 0.875$} & \scriptsize{SSIM $= 0.834$} & \scriptsize{SSIM$ = 0.821$} & \scriptsize{SSIM$ = 0.820$}\\
 &\scriptsize{PSNR $= 26.22$}  &  \scriptsize{PSNR $= 24.71$} & \scriptsize{PSNR$ = 22.91$} & \scriptsize{PSNR $= 22.51$} & \scriptsize{PSNR $= 21.96$}\\
  &\scriptsize{\textcolor{black}{CID} $= 0.276$}  &  \scriptsize{\textcolor{black}{CID} $= 0.326$} & \scriptsize{\textcolor{black}{CID}$ = 0.436$} & \scriptsize{\textcolor{black}{CID} $= 0.450$} & \scriptsize{\textcolor{black}{CID} $= 0.482$}\\
\end{tabular}
\end{center}
   \caption{Robustness of  $\mathcal{L}_2$ with outlier correspondences. The first column gives the target (top) and palette image (bottom). The remaining columns show the colour transfer results of TPS$_{rgb}^{Corr}$ when the target image is shifted horizontally by $s$ pixels, creating incorrect colour correspondences. The top row shows the shifted target superimposed on the palette image ($(T-P)_s$) and the bottom row shows our colour transfer result with corresponding SSIM, PSNR and \textcolor{black}{CID} results. }
\label{fig:RobustL2}
\end{figure*}

To evaluate the robustness of the $\mathcal{L}_2$ metric, we applied TPS$_{rgb}^{Corr}$ to images that had many false correspondences. Taking the registered palette and target images, we applied a horizontal shift of $s$ pixels to the target image. Then taking pixels at the same location in the palette and new target image as correspondences we computed the colour transfer result. Figure \ref{fig:RobustL2} shows that even when a large number of false correspondences are present, the colours in the result image are very similar to those in the palette image. Areas which have changed colour are long thin structures which no longer have many correct colour correspondences (the blue bars on the tower become green in Figure \ref{fig:RobustL2}). The structure of the target image has also been well maintained overall.

\subsubsection{Comparison to current leading techniques}

Table \ref{tab:SSIMandPSNR}  provides a quantitative evaluation of our proposed method (TPS in RGB and Lab spaces) in comparison to leading state of the art colour transfer methods \cite{Bonneel2015,Pitie2007,Hwang2014}\footnote{Results using PMLS were provided by the authors of \cite{Hwang2014}.}
(c.f. notations explained in Table \ref{tab:roadmap}). 
In terms of PSNR, SSIM and \textcolor{black}{CID}, PMLS performs slightly better in most cases, closely followed by our TPS$_{rgb}^{Corr}$ and TPS$_{lab}^{Corr}$ methods but T-tests confirm   that there is no significant quantitative difference between PMLS and each of our proposed techniques TPS$_{rgb}^{Corr}$ and TPS$_{lab}^{Corr}$
(with a 99\% confidence level). 

\begin{figure*}[!t]
\begin{center}
\begin{tabular}{cccccc}
 Target image& Palette image &  Piti\'{e} &  Bonneel  &  PMLS & TPS$_{rgb}^{Corr}$\\
\includegraphics[width = .15\linewidth, height = .1\linewidth]{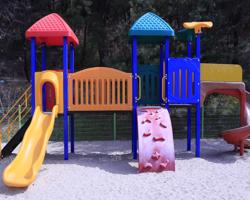} &
\includegraphics[width = .15\linewidth, height = .1\linewidth]{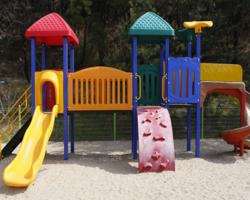} &
\includegraphics[width = .15\linewidth, height = .1\linewidth]{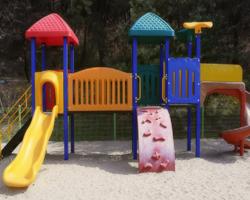}&
\includegraphics[width = .15\linewidth, height = .1\linewidth]{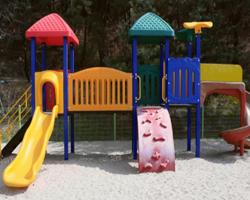}&
\includegraphics[width = .15\linewidth, height = .1\linewidth]{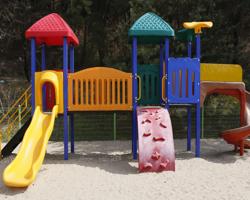} &
\includegraphics[width = .15\linewidth, height = .1\linewidth]{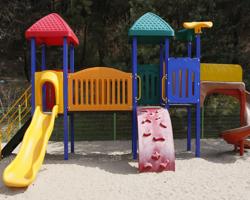} \\
\includegraphics[width = .15\linewidth, height = .09\linewidth]{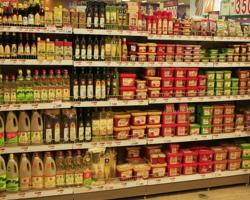} &
\includegraphics[width = .15\linewidth, height = .09\linewidth]{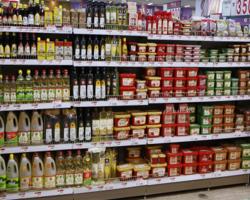} &
\includegraphics[width = .15\linewidth, height = .09\linewidth]{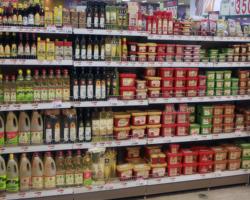}&
\includegraphics[width = .15\linewidth, height = .09\linewidth]{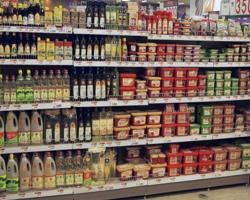}&
\includegraphics[width = .15\linewidth, height = .09\linewidth]{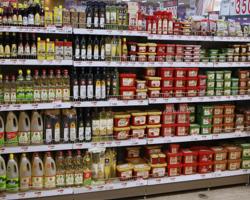} &
\includegraphics[width = .15\linewidth, height = .09\linewidth]{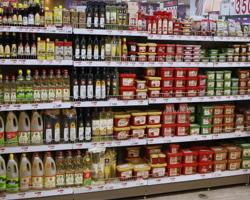} \\
\includegraphics[width = .15\linewidth, height = .09\linewidth]{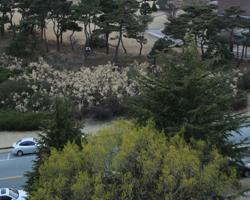} &
\includegraphics[width = .15\linewidth, height = .09\linewidth]{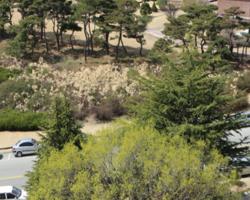} &
\includegraphics[width = .15\linewidth, height = .09\linewidth]{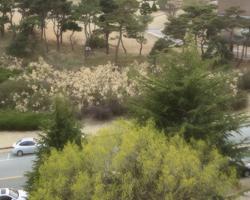}&
\includegraphics[width = .15\linewidth, height = .09\linewidth]{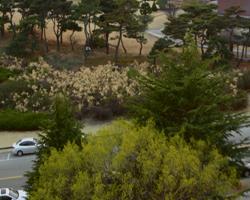}&
\includegraphics[width = .15\linewidth, height = .09\linewidth]{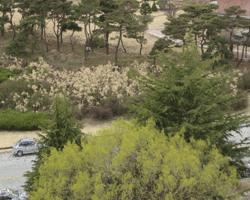} &
\includegraphics[width = .15\linewidth, height = .09\linewidth]{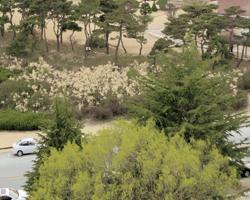} \\

\includegraphics[width = .15\linewidth, height = .09\linewidth]{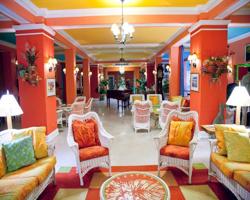} &
\includegraphics[width = .15\linewidth, height = .09\linewidth]{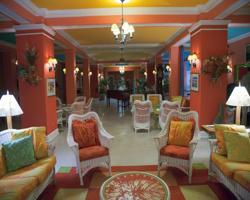} &
\includegraphics[width = .15\linewidth, height = .09\linewidth]{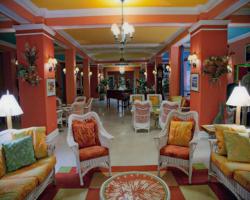}&
\includegraphics[width = .15\linewidth, height = .09\linewidth]{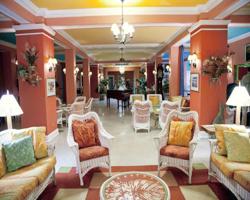}&
\includegraphics[width = .15\linewidth, height = .09\linewidth]{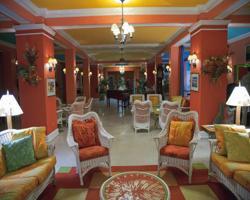} &
\includegraphics[width = .15\linewidth, height = .09\linewidth]{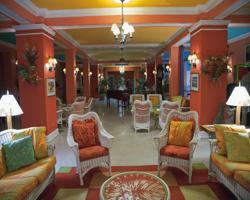} \\


\includegraphics[width = .15\linewidth, height = .1\linewidth]{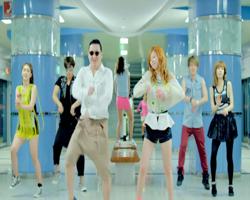} &
\includegraphics[width = .15\linewidth, height = .1\linewidth]{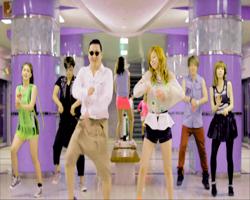} &
\includegraphics[width = .15\linewidth, height = .1\linewidth]{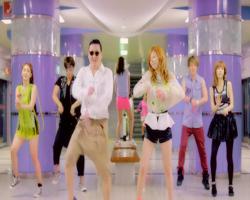}&
\includegraphics[width = .15\linewidth, height = .1\linewidth]{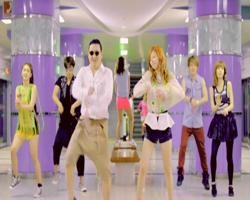}&
\includegraphics[width = .15\linewidth, height = .1\linewidth]{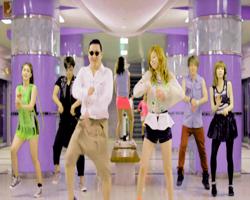} &
\includegraphics[width = .15\linewidth, height = .1\linewidth]{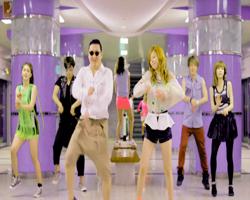} \\
\end{tabular}
\end{center}
   \caption{Results on images with similar content on the `playground', `mart', `illum', `tonal4' and `gangnam2' images. On close inspection grainy artifacts can be seen appearing in some PMLS results. For example, around the car in Row 3 Column 5 or in the top right corner of Row 2 Column 5. In comparison, the results generated by TPS$_{rgb}^{Corr}$ remain smooth (Column 6). For zoom see Fig. \ref{fig:closeUpErrors}.}
\label{fig:sameContent}
\end{figure*}

PMLS however introduces some visual artifacts  when the content in the target and palette images is not registered exactly. These artifacts can be seen around the car in row 3, column 5 of Figure \ref{fig:sameContent}.
  PMLS  is not robust to  registration errors, while  our algorithm indeed is thanks to the robust $\mathcal{L}_2$ distance. Our approach allows us to  maintain the structure of the original image and to create a smooth colour transfer result (cf. Row 3, column 6 in Fig \ref{fig:sameContent}). So to summarise while PMLS and our algorithms provide equivalent quantitative performances as measured  by PSNR, SSIM and \textcolor{black}{CID}, 
our techniques in fact provide better qualitative visual  results.

\begin{table*}[!t]
\begin{center}
\resizebox{\textwidth}{!}{
 \begin{tabular}{|l|l|l|l|l|l|l|l|l|l|l|l|l|l|l|l|}
 \hline
   \multicolumn{1}{|c|}{}& \multicolumn{5}{|c|}{PSNR} &\multicolumn{5}{|c|}{SSIM}&\multicolumn{5}{|c|}{\textcolor{black}{CID}} \\ \hline
   & {Piti\'{e}}&{ \footnotesize{Bonneel} }&{PMLS}& \tiny{TPS$_{rgb}^{Corr}$ } & \tiny{TPS$_{lab}^{Corr}$ }  & {Piti\'{e} }&\footnotesize{ Bonneel } &{PMLS}& \tiny{TPS$_{rgb}^{Corr}$ } & \tiny{TPS$_{lab}^{Corr}$ }  & {Piti\'{e}}&\footnotesize{ Bonneel }&{PMLS}& \tiny{TPS$_{rgb}^{Corr}$ } & \tiny{TPS$_{lab}^{Corr}$ } \\ \hline
    building & 20.50 & 12.32 & \textcolor{red}{22.63} & \textcolor{blue}{20.50} & \textcolor{OliveGreen}{22.51} & 0.807 & 0.675 & \textcolor{red}{0.865} & \textcolor{blue}{0.862} & \textcolor{OliveGreen}{0.864}  & 0.377  & 0.383 & \textcolor{red}{0.228} & \textcolor{OliveGreen}{0.249} & \textcolor{blue}{0.271} \\ \hline
   flower1  & 24.02 & 18.42 & \textcolor{red}{26.98} & \textcolor{OliveGreen}{26.86} & \textcolor{blue}{26.85} & 0.908 & \textcolor{blue}{0.822} & \textcolor{red}{0.967}  & \textcolor{OliveGreen}{0.966} & 0.961  & 0.395 & 0.488 & \textcolor{red}{0.163} & \textcolor{OliveGreen}{0.174} & \textcolor{blue}{0.179}\\ \hline
   flower2   & 25.32 & 21.26 & \textcolor{blue}{25.76} & \textcolor{OliveGreen}{25.77} & \textcolor{red}{25.92} & 0.900 & 0.836 & \textcolor{red}{0.928}  & \textcolor{OliveGreen}{0.927} & \textcolor{blue}{0.924} & 0.348 & 0.399 & \textcolor{red}{0.245} & \textcolor{OliveGreen}{0.266} & \textcolor{blue}{0.275} \\ \hline
   gangnam1  & 24.61 & 23.86 & \textcolor{red}{35.74} & \textcolor{blue}{35.37} & \textcolor{OliveGreen}{35.70} & 0.899 & 0.908 & \textcolor{red}{0.992}  & \textcolor{OliveGreen}{0.990} & \textcolor{blue}{0.985}  & 0.237 & 0.233 & \textcolor{red}{0.040} & \textcolor{OliveGreen}{0.048} & \textcolor{blue}{0.050} \\ \hline
   gangnam2  & 26.59 & 26.82 & \textcolor{red}{36.58} & \textcolor{OliveGreen}{35.55} & \textcolor{blue}{35.51}  & 0.918 & 0.928 & \textcolor{red}{0.993}  & \textcolor{OliveGreen}{0.986} & \textcolor{blue}{0.984} & 0.250 & 0.247 & \textcolor{red}{0.039} & \textcolor{OliveGreen}{0.089} & \textcolor{blue}{0.093}\\ \hline
   gangnam3  & 22.23 & 19.69 & \textcolor{red}{35.02} & \textcolor{OliveGreen}{33.29} & \textcolor{blue}{33.10} & 0.877 & 0.816 & \textcolor{red}{0.991}  & \textcolor{OliveGreen}{0.980} & \textcolor{blue}{0.971}  & 0.442 & 0.493 & \textcolor{red}{0.108} & \textcolor{OliveGreen}{0.193} & \textcolor{blue}{0.214}\\ \hline
   illum  & 19.89 & 14.34 & \textcolor{red}{20.17} & \textcolor{blue}{19.08} & \textcolor{OliveGreen}{19.84} & 0.632 & 0.527 & \textcolor{OliveGreen}{0.649}  & \textcolor{blue}{0.648} &  \textcolor{red}{0.650} & \textcolor{red}{0.390} & 0.411 & \textcolor{OliveGreen}{0.390} & 0.397 &\textcolor{blue}{0.397}\\ \hline
   mart &  22.71 & 22.15 & \textcolor{OliveGreen}{24.74} & \textcolor{blue}{24.45} & \textcolor{red}{24.92} & 0.906 & 0.901 & \textcolor{red}{0.957}  & \textcolor{OliveGreen}{0.956} & 0.955  & 0.513 & 0.520 & \textcolor{red}{0.219} & \textcolor{OliveGreen}{0.225} & \textcolor{blue}{0.258}\\ \hline
   playground  & 27.38 & 25.96 & \textcolor{OliveGreen}{27.84} & \textcolor{blue}{27.65} & \textcolor{red}{27.91} & 0.916 & 0.900 & \textcolor{red}{0.940}  & \textcolor{OliveGreen}{0.939} & \textcolor{blue}{0.938}  & 0.378  & 0.480  & \textcolor{red}{0.238} & \textcolor{OliveGreen}{0.254} & \textcolor{blue}{0.269}\\ \hline
   sculpture  & 29.85 & 22.57 & \textcolor{blue}{32.06} & \textcolor{OliveGreen}{32.07} & \textcolor{red}{32.10} & 0.942 & 0.873 & \textcolor{blue}{0.971}  & \textcolor{red}{0.972} & \textcolor{OliveGreen}{0.971}  & 0.227 & 0.390 & \textcolor{red}{0.137} & \textcolor{OliveGreen}{0.143} & \textcolor{blue}{0.161}\\ \hline
   tonal1 &  28.55 & 17.87 & \textcolor{OliveGreen}{37.22} & \textcolor{red}{37.33} & \textcolor{blue}{37.19} & 0.940 & 0.852 & \textcolor{red}{0.988}  & \textcolor{OliveGreen}{0.987} & \textcolor{blue}{0.987}  & 0.349 & 0.357 & \textcolor{red}{0.101} & \textcolor{OliveGreen}{0.110} & \textcolor{blue}{0.116}\\ \hline
   tonal2 & 27.88 & 23.00 & \textcolor{red}{31.51} & \textcolor{OliveGreen}{31.36} & \textcolor{blue}{31.33} & 0.968 & 0.948 & \textcolor{red}{0.987}  & \textcolor{OliveGreen}{0.986} & \textcolor{blue}{0.985}  & 0.294 & 0.292 & \textcolor{red}{0.128} & \textcolor{OliveGreen}{0.145} & \textcolor{blue}{0.152}\\ \hline
   tonal3 & 29.37 & 16.90 & \textcolor{OliveGreen}{36.25} & \textcolor{red}{36.65} & \textcolor{blue}{36.23} & 0.961 & 0.865 & \textcolor{OliveGreen}{0.992}  & \textcolor{red}{0.992} &  \textcolor{blue}{0.991}  & 0.306 & 0.418 & \textcolor{red}{0.079} & \textcolor{OliveGreen}{0.081} & \textcolor{blue}{0.083}\\ \hline
   tonal4  & 28.57 & 14.80 & \textcolor{red}{34.52} & \textcolor{blue}{34.34} &  \textcolor{OliveGreen}{34.44} & 0.943 & 0.812 & \textcolor{OliveGreen}{0.983}  & \textcolor{red}{0.983} &  \textcolor{blue}{0.983}  & 0.262 & 0.302 & \textcolor{OliveGreen}{0.108} & \textcolor{red}{0.107} & \textcolor{blue}{0.110} \\ \hline
   tonal5  & 30.20 & 21.08 & \textcolor{red}{35.26} & \textcolor{blue}{34.30} &  \textcolor{red}{34.96} & 0.965 & 0.911 & \textcolor{blue}{0.986} & \textcolor{OliveGreen}{0.985} &  \textcolor{blue}{0.984}  & 0.185 & 0.248 & \textcolor{red}{0.091}  & \textcolor{OliveGreen}{0.092} & \textcolor{blue}{0.094}\\ \hline
   
   $\mu$ & 25.85 & 20.07 & \textcolor{red}{30.82} & \textcolor{blue}{30.30} & \textcolor{OliveGreen}{30.56}  & 0.899 & 0.838 & \textcolor{red}{0.946} & \textcolor{OliveGreen}{0.944} & \textcolor{blue}{0.942} & 0.330 & 0.377 & \textcolor{red}{0.154} & \textcolor{OliveGreen}{0.172} & \textcolor{blue}{0.181} \\ \hline
 $SE$ & 0.9 & 1.1 & 1.5 & 1.5 & 1.4 & 0.02 &  0.03 & 0.02 & 0.02 & 0.02  & 0.023 & 0.024 & 0.024 & 0.024 & 0.025 \\ \hline 
 \end{tabular}}
    \caption{Comparison of our algorithm TPS$_{rgb}^{Corr}$  and TPS$_{lab}^{Corr}$  against state of the art techniques. SSIM, PSNR and \textcolor{black}{CID} results for colour transfer techniques on images with similar content: Highest PSNR and SSIM values, and lowest \textcolor{black}{CID} values, indicate the best results(best in red, second best in green, third best in blue). The overall mean score $\mu$  and its standard error $SE$ for each method are also given.}
    \label{tab:SSIMandPSNR}
\end{center}
\end{table*}

\begin{figure}[!h]
\begin{center}
\begin{tabular}{ccc}
 Target image &  PMLS &  TPS$_{rgb}^{Corr}$\\
\includegraphics[width = .3\linewidth, height = .25\linewidth]{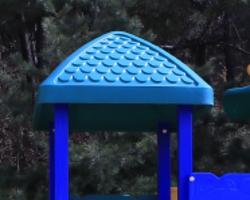}&%

\includegraphics[width = .3\linewidth, height = .25\linewidth]{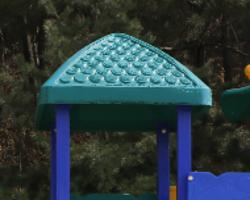} &
\includegraphics[width = .3\linewidth, height = .25\linewidth]{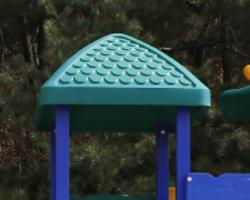}\\

\includegraphics[width = .3\linewidth, height = .25\linewidth]{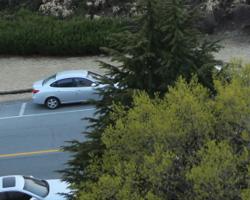}&

\includegraphics[width = .3\linewidth, height = .25\linewidth]{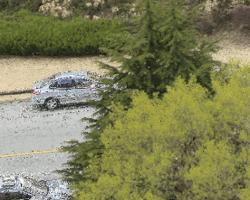} &
\includegraphics[width = .3\linewidth, height = .25\linewidth]{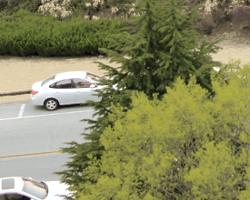}\\

\includegraphics[width = .3\linewidth, height = .25\linewidth]{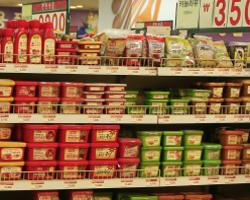}&

\includegraphics[width = .3\linewidth, height = .25\linewidth]{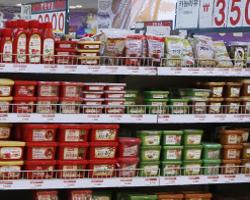} &
\includegraphics[width = .3\linewidth, height = .25\linewidth]{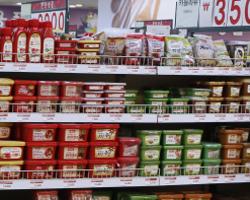}\\
\end{tabular}
\end{center}
   \caption{A close up look at some of the errors generated using PMLS \cite{Hwang2014} algorithm in comparison to our smooth result with TPS$_{rgb}^{Corr}$.  }
\label{fig:closeUpErrors}
\end{figure}

\subsection{Images with different content $P\neq T$}
\label{sec:different:content}

\begingroup
\setlength{\tabcolsep}{3pt} 
\begin{figure*}[!t]
\begin{center}
\begin{tabular}{cccccc}
  &  &  &   &   &  \\
\includegraphics[width = .15\linewidth, height = .1\linewidth]{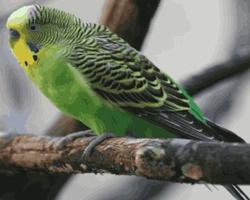} &
\includegraphics[width = .15\linewidth, height = .1\linewidth]{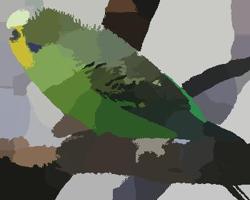} &
\includegraphics[width = .15\linewidth, height = .1\linewidth]{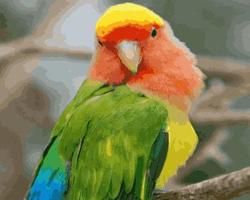} &
\includegraphics[width = .15\linewidth, height = .1\linewidth]{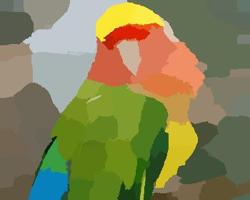} &
\includegraphics[width = .15\linewidth, height = .1\linewidth]{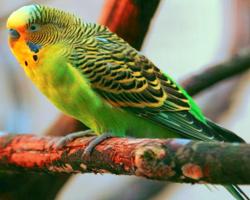}&
\includegraphics[width = .15\linewidth, height = .1\linewidth]{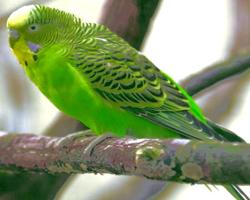} \\
\includegraphics[width = .15\linewidth, height = .1\linewidth]{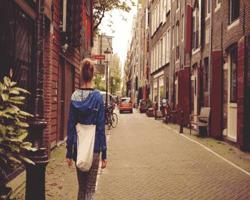} &
\includegraphics[width = .15\linewidth, height = .1\linewidth]{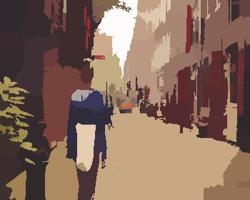} &
\includegraphics[width = .15\linewidth, height = .1\linewidth]{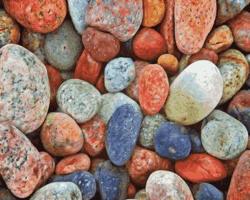} &

\includegraphics[width = .15\linewidth, height = .1\linewidth]{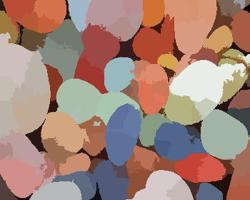} &
\includegraphics[width = .15\linewidth, height = .1\linewidth]{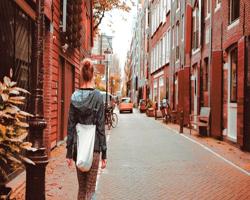}&
\includegraphics[width = .15\linewidth, height = .1\linewidth]{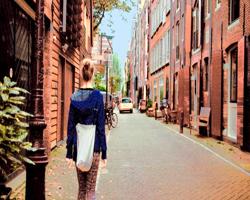} \\

\includegraphics[width = .15\linewidth, height = .1\linewidth]{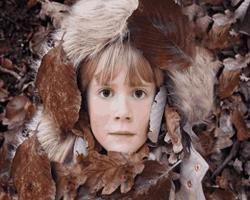} &
\includegraphics[width = .15\linewidth, height = .1\linewidth]{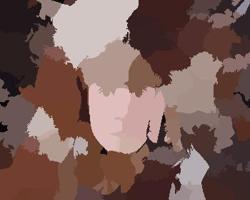} &

\includegraphics[width = .15\linewidth, height = .1\linewidth]{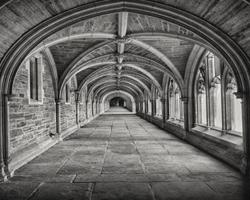} &

\includegraphics[width = .15\linewidth, height = .1\linewidth]{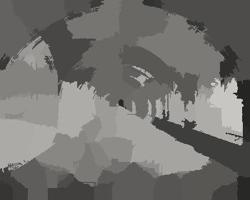} &
\includegraphics[width = .15\linewidth, height = .1\linewidth]{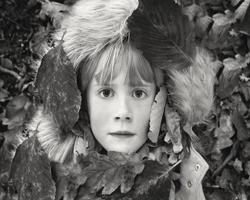}&
\includegraphics[width = .15\linewidth, height = .1\linewidth]{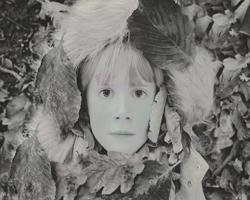} \\
K-means& Mean Shift & K-means & Mean Shift & K-means & Mean Shift \\ 
Target Clusters & Target Clusters & Palette clusters & Palette Clusters & Result TPS$_{rgb}^{KM}$
  & Result TPS$_{rgb}^{MS}$ \\
\end{tabular}
\end{center}
   \caption{Comparison between K-means clustering and Mean Shift clustering when setting the Gaussian centres $\{\mu^{(t)}\}$ of $p_t$ and $p_p$. Columns 1 to 4 show the target and palette images recoloured with the $K$ cluster centres found using either K-means or Mean Shift. Columns 5 and 6 shows the results obtained by TPS$_{rgb}$ using the clusters determined by each algorithm.  }
\label{fig:KmeansvsMeanShift}
\end{figure*}
\endgroup

First we compare two methods for estimating the Gaussian means $\{ \mu^{(k)}\}$ - K-means and Mean Shift. In Figure \ref{fig:KmeansvsMeanShift} we present the target and palette images recoloured using the $K$ cluster centres found using these techniques. The cluster centres generated using K-means are evenly spaced throughout the colour distribution of the images and the recoloured images look very similar to the original target and palette. On the other hand, the Mean Shift algorithm takes pixel colour and position into account and the cluster centres therefore depend on the structure of the image. We found that setting the Gaussian mixture means to be the K-means cluster centres gave better results than the Mean Shift clusters
 as seen in Figure \ref{fig:KmeansvsMeanShift}. Therefore we present results obtained using the K-means clustering technique in the rest of this section.

We compare our algorithm with other colour transfer techniques \cite{Bonneel2015,Ferradans2013,Pitie2007} applied to images of different content and  without correspondences.
In the case of Ferradans et al's results, all images were generated using the parameters $\lambda_X = \lambda_Y = 10^{-3} $ and $\kappa = (0.1,1,0.1,1)$ \cite{Ferradans2013}. 
 Figure \ref{fig:DiffContent} shows that  Bonneel and Ferradans  methods create blocky artifacts in the result image gradient in some cases (Row 2,4,5). On the other hand, the added constraint in Piti\'{e} algorithm which enforces a smooth image gradient ensures that these errors do not appear in their results, creating images that are more visually pleasing. Similarly, the results of our algorithm produce results that match the colours in the palette image well, while still maintaining a smooth image gradient.

 Other basis functions, which are more non-linear (see Table \ref{tab:Times}), have a tendency to fall into local minima when estimating $\theta$ in our framework when no correspondences are available. Therefore choosing the best parameters (cf. Table \ref{tab:paramChosen}) that create good results for every image is quite difficult. However, when the estimate was the global minimum, the results were very similar to TPS for both the Lab and RGB colour spaces\footnote{See supplementary material for image results. }.

\begingroup
\setlength{\tabcolsep}{1pt} 
\begin{figure*}[!t]
\begin{center}
\begin{tabular}{cccccc}
 Target  &  Palette  &  Bonneel &  Ferradans &  Piti\'{e}  &   TPS$_{rgb}^{KM}$\\

\includegraphics[width = .15\linewidth, height = .09\linewidth]{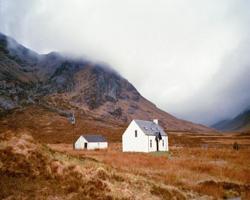} &
\includegraphics[width = .15\linewidth, height = .09\linewidth]{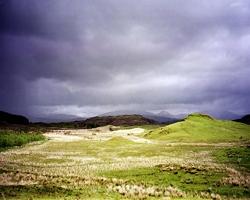} &

\includegraphics[width = .15\linewidth, height = .09\linewidth]{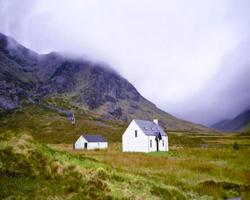} &

\includegraphics[width = .15\linewidth, height = .09\linewidth]{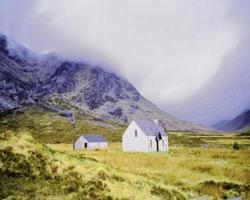} &
\includegraphics[width = .15\linewidth, height = .09\linewidth]{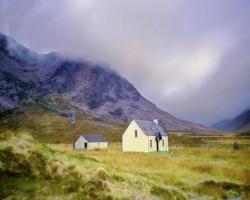}&
\includegraphics[width = .15\linewidth, height = .09\linewidth]{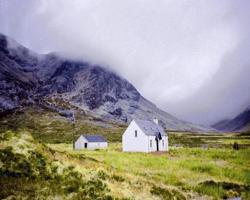} \\

\includegraphics[width = .15\linewidth, height = .09\linewidth]{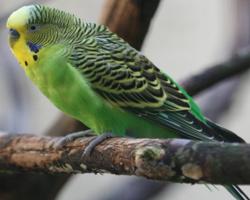} &
\includegraphics[width = .15\linewidth, height = .09\linewidth]{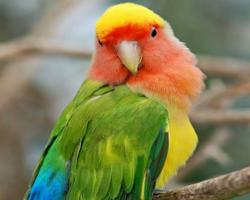} &
\includegraphics[width = .15\linewidth, height = .09\linewidth]{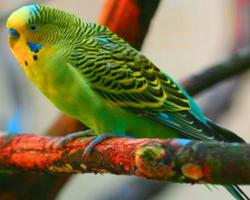} &

\includegraphics[width = .15\linewidth, height = .09\linewidth]{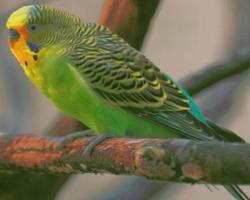} &
\includegraphics[width = .15\linewidth, height = .09\linewidth]{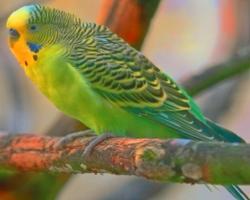}&
\includegraphics[width = .15\linewidth, height = .09\linewidth]{results/DifferentContent/L2/parrot-1parrot-2_KMeans_iter_10000_ann_5_beta_3e-06_fullCostmg_resize.jpg} \\
\includegraphics[width = .15\linewidth, height = .09\linewidth]{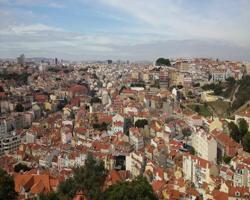} &
\includegraphics[width = .15\linewidth, height = .09\linewidth]{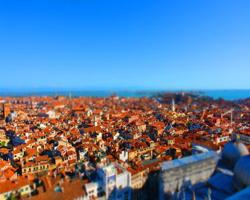} &
\includegraphics[width = .15\linewidth, height = .09\linewidth]{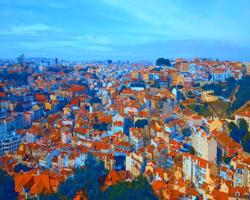} &
\includegraphics[width = .15\linewidth, height = .09\linewidth]{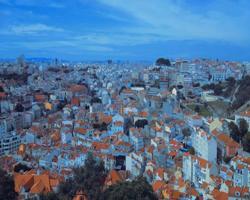} &
\includegraphics[width = .15\linewidth, height = .09\linewidth]{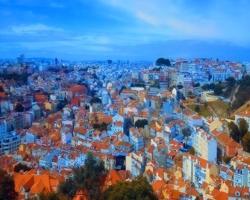}&
\includegraphics[width = .15\linewidth, height = .09\linewidth]{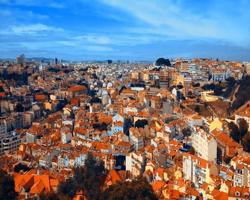}

 \\

\includegraphics[width = .15\linewidth, height = .1\linewidth]{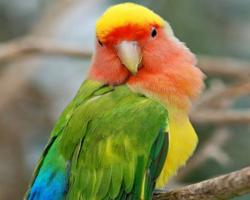} &
\includegraphics[width = .15\linewidth, height = .1\linewidth]{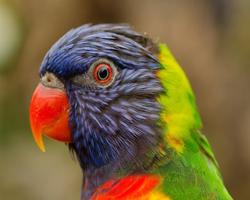} &
\includegraphics[width = .15\linewidth, height = .1\linewidth]{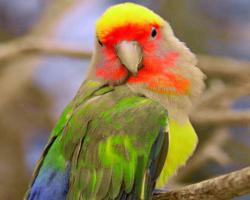} &

\includegraphics[width = .15\linewidth, height = .1\linewidth]{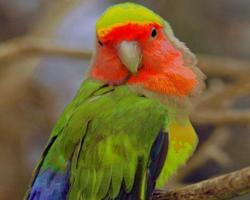} &
\includegraphics[width = .15\linewidth, height = .1\linewidth]{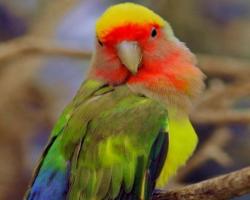}&
\includegraphics[width = .15\linewidth, height = .1\linewidth]{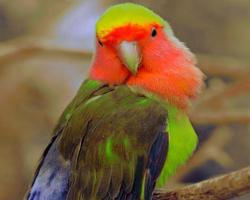}
\\
\includegraphics[width = .15\linewidth, height = .1\linewidth]{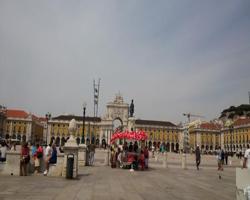} &
\includegraphics[width = .15\linewidth, height = .1\linewidth]{results/DifferentContent/Palette/parrot-7_resizemg_resize.jpg} &
\includegraphics[width = .15\linewidth, height = .1\linewidth]{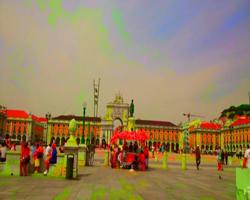} &
\includegraphics[width = .15\linewidth, height = .1\linewidth]{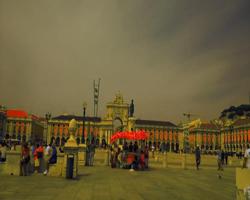} &
\includegraphics[width = .15\linewidth, height = .1\linewidth]{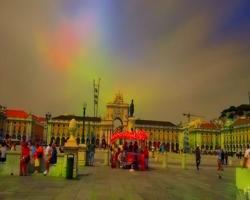}&
\includegraphics[width = .15\linewidth, height = .1\linewidth]{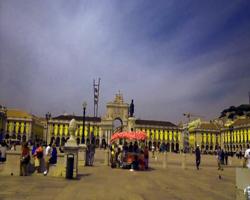}\\

\includegraphics[width = .15\linewidth, height = .1\linewidth]{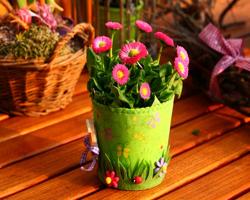} &
\includegraphics[width = .15\linewidth, height = .1\linewidth]{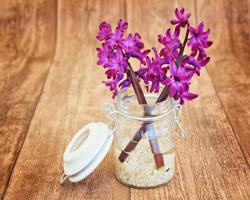} &
\includegraphics[width = .15\linewidth, height = .1\linewidth]{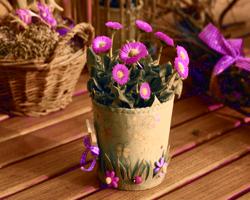} &
\includegraphics[width = .15\linewidth, height = .1\linewidth]{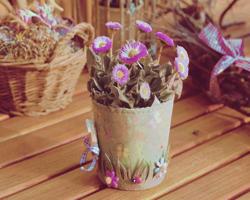} &
\includegraphics[width = .15\linewidth, height = .1\linewidth]{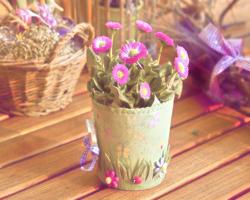}&
\includegraphics[width = .15\linewidth, height = .1\linewidth]{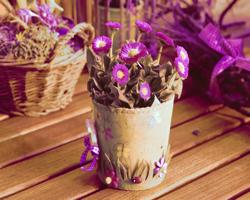}\\

\includegraphics[width = .15\linewidth, height = .1\linewidth]{results/DifferentContent/Target/hyacinthmg_resize.jpg} &
\includegraphics[width = .15\linewidth, height = .1\linewidth]{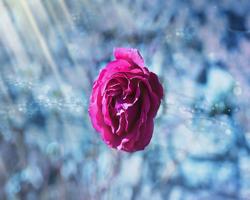} &
\includegraphics[width = .15\linewidth, height = .1\linewidth]{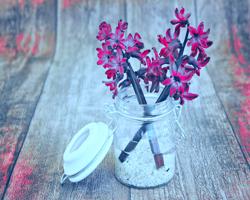} &
\includegraphics[width = .15\linewidth, height = .1\linewidth]{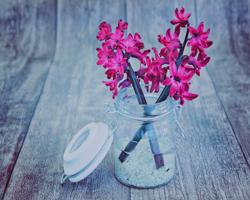} &
\includegraphics[width = .15\linewidth, height = .1\linewidth]{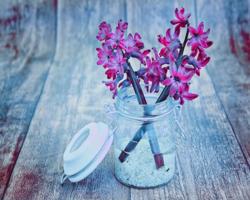}&
\includegraphics[width = .15\linewidth, height = .1\linewidth]{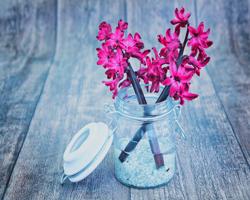}\\

\includegraphics[width = .15\linewidth, height = .1\linewidth]{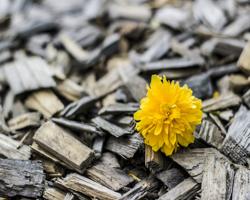} &
\includegraphics[width = .15\linewidth, height = .1\linewidth]{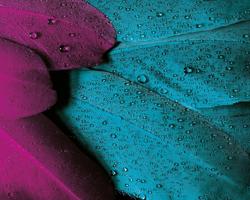} &
\includegraphics[width = .15\linewidth, height = .1\linewidth]{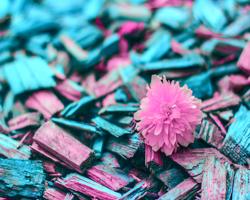} &
\includegraphics[width = .15\linewidth, height = .1\linewidth]{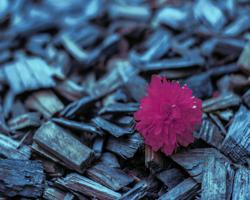} &
\includegraphics[width = .15\linewidth, height = .1\linewidth]{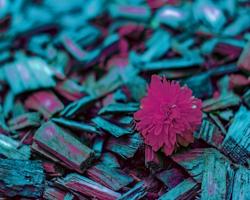}&
\includegraphics[width = .15\linewidth, height = .1\linewidth]{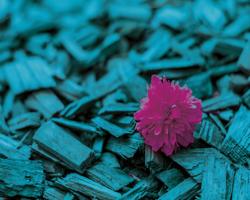}\\

\includegraphics[width = .15\linewidth, height = .1\linewidth]{results/DifferentContent/Target/pinkmg_resize.jpg} &
\includegraphics[width = .15\linewidth, height = .1\linewidth]{results/DifferentContent/Target/yellowmg_resize.jpg} &
\includegraphics[width = .15\linewidth, height = .1\linewidth]{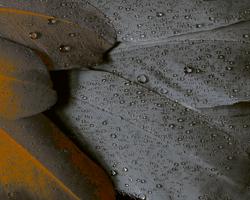} &
\includegraphics[width = .15\linewidth, height = .1\linewidth]{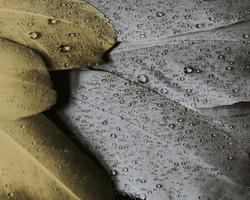} &
\includegraphics[width = .15\linewidth, height = .1\linewidth]{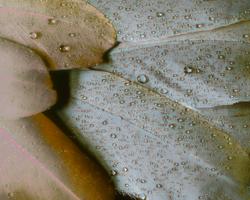}&
\includegraphics[width = .15\linewidth, height = .1\linewidth]{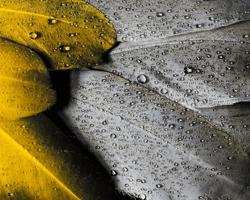}\\

\end{tabular}
\end{center}
   \caption{Some results on images with different content which were presented to participants in the user study. }
\label{fig:DiffContent}
\end{figure*}
\endgroup

\subsection{Qualitative assessment}
\label{sec:qualitative:user:study}

\textcolor{black}{Colour transfer methods have also been evaluated using two subjective user studies, each with 20 participants  evaluating 53 sets of images. In each experiment the users were asked to choose the colour transfer result that they thought was the most successful.} Out of these 53 sets of images, 38 of them had a palette and target image with different content (no correspondences available), and 15 of them contained a palette and target image with the same content (correspondences available and used). These 15 images were taken from the dataset provided by Hwang et al. (cf. paragraph \ref{sec:same:content}).  For both user studies, each user evaluated the results individually and the display properties and indoor conditions were kept constant. The order in which the image sets appeared was randomised for each user and a short trial run took place before each user study to ensure the users adapted to the task at hand.

\subsubsection{User Study I}
In our first experiment, each participant was presented with six images at a time - a target image, a palette image, and four result images generated using different colour transfer techniques. They then had 20 seconds to view the images (presented simultaneously side by side) and decide which result image was the best. For target and palette images with similar content the four methods compared were TPS$_{rgb}$, PMLS, Piti\'{e} and Bonneel. The total number of times each method was chosen can be seen in Table \ref{tab:userResultsI}. \textcolor{black}{Converting these votes to proportions, we computed confidence intervals on each proportion and compared methods by determining if their confidence intervals overlap. We found that TPS$_{rgb}$ was the best method in terms of votes however the overlapping confidence intervals indicate that TPS$_{rgb}$ and PMLS are not statistically significantly different.} Hence these two methods can be thought of as performing equally well and both being superior to  Piti\'{e} and Bonneel methods. Similarly for images with different content we compared TPS$_{rgb}$, Ferradans, Piti\'{e} and Bonneel. The total number of times each method was chosen can be seen in Table \ref{tab:userResultsI}. \textcolor{black}{Again, by comparing the confidence intervals on the proportions} we determined that TPS$_{rgb}$ performs better than both Ferradans and Bonneel, however there is no statistically significant difference between  TPS$_{rgb}$ and Piti\'{e}. 

\subsubsection{User Study II}
\textcolor{black}{For our second experiment we used a 2 alternate forced choice (2AFC) comparison. Each participant was presented with two images at a time - a target image, a palette image, and two different result images. They then had 15 seconds to view the images and pick the best result. For target and palette images with similar content we compared two methods - TPS$_{rgb}$ and PMLS, and the votes are given in Table \ref{tab:userResultsII}. Again we converted these votes to proportions and computed the confidence interval on each, and determined that there was no significant difference between the two. We also computed the Thurstone Case V analysis on the votes, which is commonly used to analyse 2AFC results \cite{morovivc2008}. We computed the preference score and confidence intervals for each  method (Table \ref{tab:userResultsII}) and found that there is a significant difference between them according to this analysis. Similarly, for images with different content we compared TPS$_{rgb}$ and  Piti\'{e}, the results of which can be seen in Table \ref{tab:userResultsII}. In this case both methods were chosen an equal number of times, indicating that there is no perceivable difference between TPS$_{rgb}$ and  Piti\'{e}.}

\begin{table}[!h]
\begin{center}
 \begin{tabular}{|c||c|c|c|}
  \multicolumn{4}{c}{User Study I }\\
   \hline

    \multicolumn{4}{|c|}{Similar Content $P\simeq T$}\\ \hline
    Method & Votes & \textcolor{black}{Prop} & \textcolor{black}{CI on Prop} \\ \hline \hline
    Bonneel & 38 & 0.126 & (0.089, 0.164)\\ \hline
    Pitie & \textcolor{blue}{ 63} &  \textcolor{blue}{0.210} & (0.164, 0.256) \\ \hline
    PMLS &  \textcolor{OliveGreen}{98} &  \textcolor{OliveGreen}{0.327} & (0.274, 0.380) \\ \hline
    TPS$_{rgb}^{Corr}$ &  \textcolor{red}{101} &  \textcolor{red}{0.337} & (0.283, 0.390) \\ \hline
    \hline

     \multicolumn{4}{|c|}{Different Content  $P\neq T$}\\ \hline
    Method & Votes & \textcolor{black}{Prop} & \textcolor{black}{CI on Prop} \\ \hline \hline
    Bonneel & 163 & 0.214 & (0.185, 0.244)\\ \hline
    Pitie & \textcolor{OliveGreen}{211} & \textcolor{OliveGreen}{0.278} & (0.246, 0.310) \\ \hline
    Ferradans & 152 & 0.20 & (0.172, 0.228) \\ \hline
    TPS$_{rgb}^{KM}$ & \textcolor{red}{234} & \textcolor{red}{0.308} & (0.275, 0.341) \\ \hline

     \end{tabular}
    \caption{Number of votes given to each method by participants in our first perceptual  study (best indicated in red, second best in green, third best in blue), their corresponding proportion and confidence interval.}
    \label{tab:userResultsI}
    \end{center}
\end{table}

\begin{table}[!h]
\begin{center}
 \begin{tabular}{|c|c|c|c|c|c|}
 \multicolumn{6}{c}{\textcolor{black}{User Study II }}\\
 \hline
    \multicolumn{6}{|c|}{Similar Content $P\simeq T$}\\ \hline
    Method & Votes & Prop & CI on Prop & Score & CI on score  \\ \hline \hline
     PMLS & 134 & 0.447 & (0.390, 0.503)& -0.190 & (0.303, -0.076)  \\ \hline
     TPS$_{rgb}^{Corr}$ & \textcolor{red}{166} & \textcolor{red}{0.553} & (0.497, 0.610) & 0.190 & (0.077, 0.303)  \\ \hline
     
     \multicolumn{6}{|c|}{Different Content  $P\neq T$}\\ \hline
     Method & Votes & Prop & CI on Prop & Score & CI on score \\ \hline \hline
      Piti\'{e} & \textcolor{red}{380} & \textcolor{red}{0.5} & (0.443, 0.557) & 0 & (-0.1132, 0.1132)\\ \hline
      TPS$_{rgb}^{KM}$ & \textcolor{red}{380} & \textcolor{red}{0.5} & (0.443, 0.557) & 0 & (-0.1132, 0.1132)\\ \hline

     \end{tabular}
    \caption{Number of votes given to each method by participants in our second perceptual study (best indicated in red, second best in green, third best in blue), corresponding proportion and confidence interval, Thurstone score and associated confidence interval.}
    \label{tab:userResultsII}
    \end{center}
\end{table}

\subsection{Computation Time}
\label{sec:computation:time}

It is important to provide  timely feedback  for artists and amateurs alike when  recolouring image and video materials. 
The recolouring step is highly parallelisable, allowing the recolouring of video content at a high speed. 
In terms of computation, our algorithm is split into three parts: the clustering step, the estimation step of $\theta$ and the recolouring step $x\rightarrow \phi(x,\theta)$. As K-means can be quite a time consuming clustering technique, we investigated some fast quantisation methods including those provided by Matlab and the GNU Image Manipulation Program (GIMP)\cite{GIMP}. We found that using Matlab's minimum variance quantisation method (MVQ)\cite{Matlabrgb2ind} provided almost identical results to K-means as well as being much faster and could be used as an alternative to K-means to speed up the clustering step (Table \ref{tab:Times}). A comparison between these techniques and K-means can be seen in Figure \ref{fig:quant}.

The estimation step takes approximately 6 seconds using non optimised code. Once $\theta$ is estimated however, it can then be used to recolour an image of any size, or be applied to recolour a video clip. It can also be stored for later usage (see paragraph \ref{sec:visual:effects}). 


The recolouring step on the other hand is highly parallelisable and can be applied independently to each pixel. The time taken to recolour a HD image for each type of radial basis function is given in Table \ref{tab:Times}. In our implementation we used OpenMP within a Matlab mex file to parallelise this step on 8 CPU threads. All times are computed on a 2.93GHz Intel CPU with 3GB of RAM with 4 cores and 8 logical processors. 
In comparison, the  GPU implementation of PMLS  takes 4.5 seconds to recolour a 1 million pixel image using an nVIDIA Quadro 4000 as reported in \cite{Hwang2014}, which is 9 times slower than our implementation with TPS. 
Similarly, Bonneel et al. report  a time of approximately 3 minutes to recolour 108 frames of a HD segmented video on an 8 core machine with their algorithm \cite{Bonneel2013}. In comparison, our algorithm would take less than 2 minutes using TPS  in the same situation.

It can also be applied to each pixel in parallel and a GPU implementation would allow videos to be recoloured extremely quickly. 

\begingroup
\setlength{\tabcolsep}{2pt}
\begin{figure}[!t]
\begin{center}
\begin{tabular}{ccccc}
\includegraphics[width = .19\linewidth, height = .15\linewidth]{results/DifferentContent/Target/scotland_housemg_resize.jpg} &
\includegraphics[width = .19\linewidth, height = .15\linewidth]{results/DifferentContent/Palette/scotland_plainmg_resize.jpg}  &
\includegraphics[width = .19\linewidth, height = .15\linewidth]{results/DifferentContent/L2/scotland_housescotland_plain_KMeans_iter_10000_ann_5_beta_3e-06_fullCostmg_resize.jpg}&
\includegraphics[width = .19\linewidth, height = .15\linewidth]{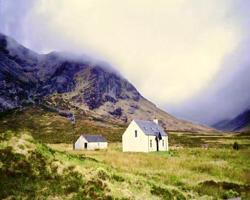}&
\includegraphics[width = .19\linewidth, height = .15\linewidth]{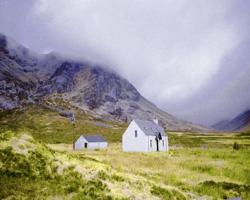}\\
\includegraphics[width = .19\linewidth, height = .15\linewidth]{results/DifferentContent/Target/parrot-1mg_resize.jpg} &
\includegraphics[width = .19\linewidth, height = .15\linewidth]{results/DifferentContent/Palette/parrot-2mg_resize.jpg}  &
\includegraphics[width = .19\linewidth, height = .15\linewidth]{results/DifferentContent/L2/parrot-1parrot-2_KMeans_iter_10000_ann_5_beta_3e-06_fullCostmg_resize.jpg}&
\includegraphics[width = .19\linewidth, height = .15\linewidth]{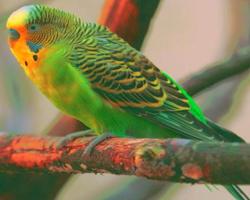}&
\includegraphics[width = .19\linewidth, height = .15\linewidth]{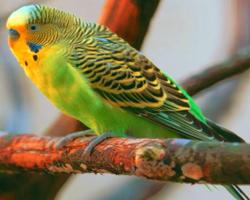}\\
\includegraphics[width = .19\linewidth, height = .15\linewidth]{results/DifferentContent/Target/lisbon1_resizemg_resize.jpg} &
\includegraphics[width = .19\linewidth, height = .15\linewidth]{results/DifferentContent/Palette/parrot-7_resizemg_resize.jpg}  &
\includegraphics[width = .19\linewidth, height = .15\linewidth]{results/DifferentContent/L2/lisbon1_resizeparrot-7_resize_KMeans_iter_10000_ann_5_beta_3e-06_fullCostmg_resize.jpg}&
\includegraphics[width = .19\linewidth, height = .15\linewidth]{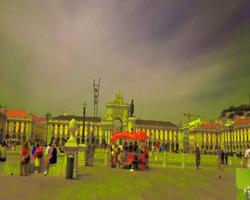}&
\includegraphics[width = .19\linewidth, height = .15\linewidth]{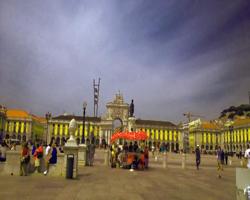}\\
Target & Palette & TPS$^{KM}_{rgb}$ & TPS$^{GIMP}_{rgb}$ & TPS$^{MVQ}_{rgb}$ \\
\end{tabular}
\end{center}
   \caption{Comparison between the results obtained using the K-means clustering technique and two faster quantisation techniques. TPS$^{GIMP}_{rgb}$ are the results obtained when the target and palette images have been clustered using GIMPs quantisation method (the median cut algorithm). TPS$^{MVQ}_{rgb}$ are the results obtained when Matlab's minimum variance quantisation algorithm is used to cluster the images. TPS$^{MVQ}_{rgb}$ gives results that are very similar to TPS$^{KM}_{rgb}$. }
\label{fig:quant}
\end{figure}
\endgroup

\begin{table}[!h]
\begin{center}

 \begin{tabular}{|l|l|}
 \multicolumn{2}{c}{\textbf{Clustering}}\\ \hline
  \textbf{Method}  & \textbf{Time} \\ \hline
    K-means & 10.4s\\ 
  
    Mean Shift  & 3.2s\\ 
    MVQ  & 0.005s\\ \hline
    
   \multicolumn{2}{c}{\textbf{Estimating $\hat{\theta}$}}\\ \hline
   \textbf{K}  & \textbf{Time} \\ \hline
    K = 50 & 6s \\ \hline

   \multicolumn{2}{c}{\textbf{Recolouring}}\\ \hline
   \textbf{RBF name}  & \textbf{Time} \\ \hline
    TPS & 1.04s\\ 
  
   Gaussian  & 3.31s \\ 
   Inverse Multiquadric  &3.16s \\ 
   Inverse Quadric  & 2.74s \\ \hline
 \end{tabular}
 \end{center}
    \caption{Computation times in seconds for each step of our algorithm for a HD (1080 $\times$ 1920) image with over 2 million pixels. (For the clustering step, the images were first downsampled to $300 \times 350$ pixels to reduce computation time). }
    \label{tab:Times}
\end{table}

%

\section{Usability for recolouring}
\label{sec:usability}

The estimated parametric transfer function computed by our algorithms  can be stored and  combined easily with other transfer functions computed using various colour 
palettes for the creation of visual effects (cf. Section \ref{sec:visual:effects}). Existing postprocessing can also be used to further improve the quality of the results (Section \ref{sec:postprocessing}).

\subsection{Mixing colour palettes}
\label{sec:visual:effects}

Once estimated, the parametric transformation can be easily applied to video content \cite{Grogan2015b,Frigo15}. 
Interpolating between two parametric transformations $\hat{\theta}^{(1)}$ and $\hat{\theta}^{(2)}$   can also be used to create interesting effects in images.

The interpolating weight can also vary over time $\gamma(t)\in [0;1]$ to create dissolve effects between colour palettes in videos, and can be applied as follows to each colour pixel $x$ at pixel spatial location $p=(p_1,p_2)$ in the sequence at time $t$ \cite{Grogan2015b}:
\begin{equation}
x(p,t) \rightarrow \phi\left(x,\gamma(t)\ \hat{\theta}^{(1)}+(1-\gamma(t))\ \hat{\theta}^{(2)} \right).
\end{equation}
We extend this idea  further using interpolation weights that vary spatially, as well as in the temporal domain, where pixels   are recoloured  as follows: 
\begin{equation}
x(p,t) \rightarrow \phi\left(x,\gamma(p,t)\ \hat{\theta}^{(1)}+(1-\gamma(p,t))\ \hat{\theta}^{(2)} \right)
\end{equation}
$\gamma(p,t)$ is a dynamic greyscale mask with values between 0 and 1. Figure \ref{fig:effects} presents some examples of results on images
\footnote{Supplementary results for   videos are provided by the authors.}. 
Moreover these effects can be extended to mixing more than 2 transformations (or palettes).
A simpler interpolation between the identity transformation $\theta^{(Id)}$ and an estimated transformation $\hat{\theta}$ with a selected colour palette can also be created:
\begin{equation}
x(p,t)\rightarrow \phi\left(x,\gamma(t)\ \hat{\theta}^{(Id)}+(1-\gamma(t))\ \hat{\theta} \right)  
\end{equation}
This gives the user the flexibility to transition smoothly from one colour mood to another.

\begingroup
\setlength{\tabcolsep}{1pt} 
\begin{figure}[!h]
\begin{center}
\begin{tabular}{ccc}
\includegraphics[width = .2\linewidth, height = .13\linewidth]{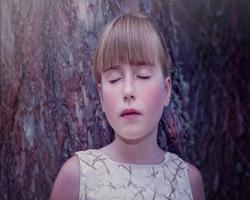} &
\includegraphics[width = .2\linewidth, height = .13\linewidth]{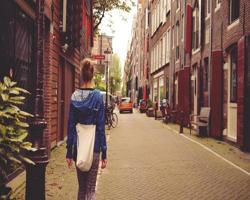} &
\multirow{2}{*}[0.33in]{\includegraphics[width = .4\linewidth, height = .27\linewidth]{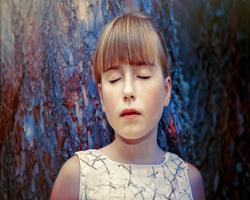}}\\ \vspace{6pt}
\includegraphics[width = .2\linewidth, height = .13\linewidth]{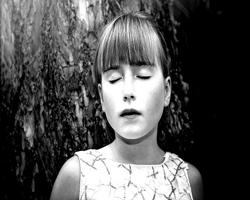} &
\includegraphics[width = .2\linewidth, height = .13\linewidth]{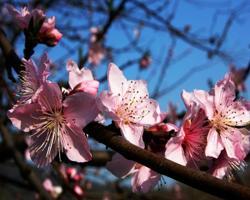}& \\


\includegraphics[width = .2\linewidth, height = .13\linewidth]{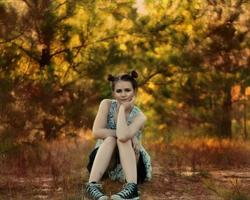} &
\includegraphics[width = .2\linewidth, height = .13\linewidth]{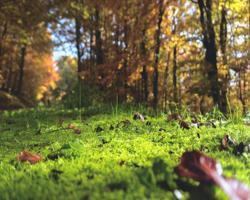} &
\multirow{2}{*}[0.33in]{\includegraphics[width = .4\linewidth, height = .27\linewidth]{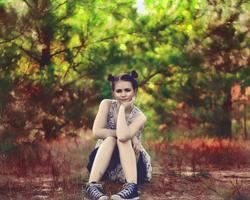}}\\ \vspace{6pt}
\includegraphics[width = .2\linewidth, height = .13\linewidth]{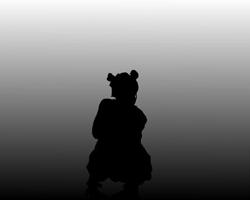} &
\includegraphics[width = .2\linewidth, height = .13\linewidth]{results/DownsizedPalettes/amsterdam_resizevsmallmg_resize.jpg}& \\ 

\includegraphics[width = .2\linewidth, height = .13\linewidth]{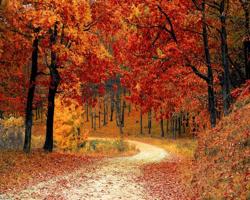} &
\includegraphics[width = .2\linewidth, height = .13\linewidth]{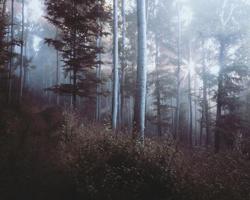} &
\multirow{2}{*}[.33in]{\includegraphics[width = .4\linewidth, height = .27\linewidth]{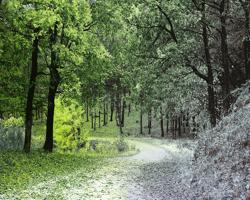}}\\\vspace{6pt} 

\includegraphics[width = .2\linewidth, height = .13\linewidth]{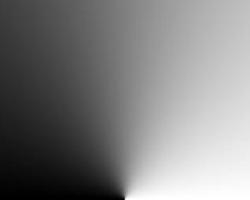} &
\includegraphics[width = .2\linewidth, height = .13\linewidth]{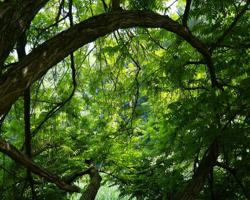}& \\ 
Target \& Mask & Palettes 1 \& 2 & Results TPS$_{rgb}^{KM}$\\ 

\end{tabular}
\end{center}
   \caption{Mask $\gamma(p)$ is used to vary the recolouring spatially. The results images were generated using the target and mask (column 1) and two palette images (column 2). Pixels in the target image whose corresponding pixel in the grey scale mask is white have been recoloured using palette 1 (top), and those whose corresponding pixel is black have been recoloured using palette 2(bottom). The remaining pixels have been recoloured using an interpolation between the two transformations. For each pixel this interpolation is determined by the value of its corresponding pixel in the grey scale mask. Images shown are public domain images sourced from pixabay.com.}
\label{fig:effects}
\end{figure}
\endgroup

\subsection{Post-processing}
\label{sec:postprocessing}

While our approach produces excellent results in general, rare saturation artifacts can occur
when many colours in the palette image lie close to the boundary of the colour space. As we do not constrain the transformed colours to lie within a specific range, some colours can potentially get transformed outside of the colour space. Currently we clamp the colour values so that they are within the desired range. Recently Oliveira et al. \cite{Oliveira2015} proposed using truncated Gaussians which could be implemented in future to overcome this problem. 
Additionally,  a post-processing step could be added to our pipeline when needed,  to constrain the smoothness of the gradient field of the recoloured image to be similar to the gradient field of the target image  as proposed by  \cite{Pitie2007} (c.f. Figure \ref{fig:grainy}).

\begingroup
\setlength{\tabcolsep}{3pt} 
\begin{figure}[!t]
\begin{center}
\begin{tabular}{ccc}
\includegraphics[width = .24\linewidth, height = .2\linewidth]{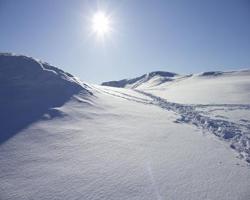} &
\includegraphics[width = .24\linewidth, height = .2\linewidth]{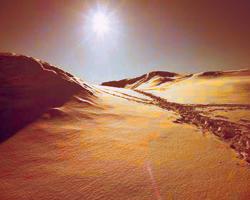} &
\includegraphics[width = .24\linewidth, height = .2\linewidth]{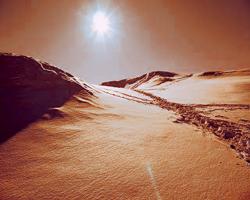}\\
\includegraphics[width = .24\linewidth, height = .2\linewidth]{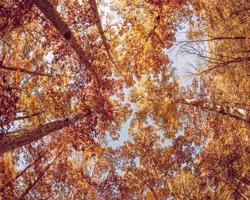} &
\includegraphics[width = .24\linewidth, height = .2\linewidth]
{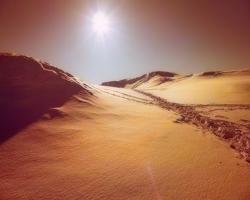} &
\includegraphics[width = .24\linewidth, height = .2\linewidth]{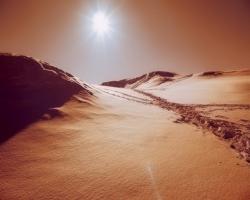} \\
Target \& Palette & Piti\'{e} et al.\cite{Pitie2005,Pitie2007} & TPS$_{rgb}^{KM}$   \\ 

\end{tabular}
\end{center}
   \caption{Example of quantisation errors enhanced by our transfer method (e.g. in the sky) as the gradient becomes stretched to match the colours in the palette image (row 1, col 3). The gradient smoothing step proposed by Piti\'{e} et al. \cite{Pitie2007} as an improvement from their ealier method without post processing \cite{Pitie2005} could be used in our pipeline to remove artifacts,   although in general this can reduce the sharpness of the image which is undesirable. Column 2: Piti\'{e} et al. \cite{Pitie2005} (before smoothing) and \cite{Pitie2007} (after smoothing); Column 2: TPS$_{rgb}^{KM}$ (before smoothing) and TPS$_{rgb}^{KM}$ (after smoothing). }
\label{fig:grainy}
\end{figure}
\endgroup

\section{Conclusion }
\label{sec:conclusion}

We have presented a new  framework  for  colour transfer that is  able to take into account correspondences when these are available.
Our  algorithms compete very well with current state of the art approaches since no 
statistical differences can be measured using  metrics such as SSIM, PSNR and \textcolor{black}{CID} with the current top techniques, and 
 visual inspection of our results show that our algorithms are more immune to occasional artifacts that can be created in the gradient field of the recoloured image.  
Our parametric formulation of the transfer function  allows for fast recolouring of images and videos. Moreover transfer functions can be stored and easily 
combined and interpolated for creating visual effects. 
 Future work will investigate  techniques to capture users' preferences by learning from exemplar transfer functions \cite{ZhichengACMTOG2016}.

\section*{Acknowledgements}
\small 
This work has been supported by a Ussher scholarship from Trinity College
Dublin (Ireland), and partially supported by EU FP7-PEOPLE-2013-
IAPP GRAISearch grant (612334).
\normalsize

\bibliographystyle{IEEEtran}
\bibliography{IPbibliography}

\begin{thebibliography}{10}
\providecommand{\url}[1]{#1}
\csname url@samestyle\endcsname
\providecommand{\newblock}{\relax}
\providecommand{\bibinfo}[2]{#2}
\providecommand{\BIBentrySTDinterwordspacing}{\spaceskip=0pt\relax}
\providecommand{\BIBentryALTinterwordstretchfactor}{4}
\providecommand{\BIBentryALTinterwordspacing}{\spaceskip=\fontdimen2\font plus
\BIBentryALTinterwordstretchfactor\fontdimen3\font minus
  \fontdimen4\font\relax}
\providecommand{\BIBforeignlanguage}[2]{{%
\expandafter\ifx\csname l@#1\endcsname\relax
\typeout{** WARNING: IEEEtran.bst: No hyphenation pattern has been}%
\typeout{** loaded for the language `#1'. Using the pattern for}%
\typeout{** the default language instead.}%
\else
\language=\csname l@#1\endcsname
\fi
#2}}
\providecommand{\BIBdecl}{\relax}
\BIBdecl

\bibitem{Grogan2015}
M.~Grogan, M.~Prasad, and R.~Dahyot, ``L2 registration for colour transfer,''
  in \emph{Proceedings of the 23rd European Signal Processing Conference
  (EUSIPCO)}, ser. EUSIPCO '15, Nice, France, 2015, pp. 2366--2370.

\bibitem{Grogan2015b}
M.~Grogan and R.~Dahyot, ``L2 registration for colour transfer in videos,'' in
  \emph{Proceedings of the 12th European Conference on Visual Media
  Production}, ser. CVMP '15.\hskip 1em plus 0.5em minus 0.4em\relax New York,
  NY, USA: ACM, 2015, pp. 16:1--16:1.

\bibitem{Lefebvre2014}
H.~S. Faridul, T.~Pouli, C.~Chamaret, J.~Stauder, A.~Tremeau, and E.~Reinhard,
  ``{A Survey of Color Mapping and its Applications},'' in \emph{Eurographics
  2014 - State of the Art Reports}, S.~Lefebvre and M.~Spagnuolo, Eds.\hskip
  1em plus 0.5em minus 0.4em\relax The Eurographics Association, 2014.

\bibitem{NaranjoICIP2000}
V.~Naranjo and A.~Albiol, ``Flicker reduction in old films,'' in
  \emph{International Conference on Image Processing (ICIP)}, 2000.

\bibitem{Reinhard2001}
E.~Reinhard, M.~Adhikhmin, B.~Gooch, and P.~Shirley, ``Color transfer between
  images,'' \emph{Computer Graphics and Applications, IEEE}, vol.~21, no.~5,
  pp. 34--41, Sep 2001.

\bibitem{Bonneel2015}
N.~Bonneel, J.~Rabin, G.~Peyré, and H.~Pfister,
  ``\BIBforeignlanguage{English}{Sliced and radon wasserstein barycenters of
  measures},'' \emph{\BIBforeignlanguage{English}{Journal of Mathematical
  Imaging and Vision}}, vol.~51, no.~1, pp. 22--45, 2015.

\bibitem{Hwang2014}
Y.~Hwang, J.-Y. Lee, I.~S. Kweon, and S.~J. Kim, ``Color transfer using
  probabilistic moving least squares,'' in \emph{Computer Vision and Pattern
  Recognition (CVPR), 2014 IEEE Conference on}, June 2014, pp. 3342--3349.

\bibitem{Ferradans2013}
S.~Ferradans, N.~Papadakis, J.~Rabin, G.~Peyré, and J.-F. Aujol,
  ``\BIBforeignlanguage{English}{Regularized discrete optimal transport},'' in
  \emph{\BIBforeignlanguage{English}{Scale Space and Variational Methods in
  Computer Vision}}, ser. Lecture Notes in Computer Science, A.~Kuijper,
  K.~Bredies, T.~Pock, and H.~Bischof, Eds.\hskip 1em plus 0.5em minus
  0.4em\relax Springer Berlin Heidelberg, 2013, vol. 7893, pp. 428--439.

\bibitem{Pitie2007}
F.~Piti{\'e}, A.~C. Kokaram, and R.~Dahyot, ``Automated colour grading using
  colour distribution transfer,'' \emph{Comput. Vis. Image Underst.}, vol. 107,
  no. 1-2, pp. 123--137, Jul. 2007.

\bibitem{VillaniBook2009}
C.~Villani, \emph{Optimal transport : old and new}, ser. Grundlehren der
  mathematischen Wissenschaften.\hskip 1em plus 0.5em minus 0.4em\relax Berlin:
  Springer, 2009.

\bibitem{Villani2003}
------, \emph{Topics in Optimal Transportation}.\hskip 1em plus 0.5em minus
  0.4em\relax American Mathematical Society, 2003, vol.~58.

\bibitem{Pitie2005}
F.~Pitie, A.~Kokaram, and R.~Dahyot, ``N-dimensional probability density
  function transfer and its application to color transfer,'' in \emph{Computer
  Vision, 2005. ICCV 2005. Tenth IEEE International Conference on}, vol.~2, Oct
  2005, pp. 1434--1439 Vol. 2.

\bibitem{Neumann2005}
L.~Neumann and A.~Neumann, ``Color style transfer techniques using hue,
  lightness and saturation histogram matching,'' in \emph{Proceedings of the
  First Eurographics Conference on Computational Aesthetics in Graphics,
  Visualization and Imaging}, ser. Computational Aesthetics'05.\hskip 1em plus
  0.5em minus 0.4em\relax Aire-la-Ville, Switzerland, Switzerland: Eurographics
  Association, 2005, pp. 111--122.

\bibitem{Papadakis2011}
N.~Papadakis, E.~Provenzi, and V.~Caselles, ``A variational model for histogram
  transfer of color images,'' \emph{Image Processing, IEEE Transactions on},
  vol.~20, no.~6, pp. 1682--1695, June 2011.

\bibitem{Pouli2011}
T.~Pouli and E.~Reinhard, ``Progressive color transfer for images of arbitrary
  dynamic range,'' \emph{Computers \& Graphics}, vol.~35, no.~1, pp. 67 -- 80,
  2011, extended Papers from Non-Photorealistic Animation and Rendering (NPAR)
  2010.

\bibitem{Freedman2010}
D.~Freedman and P.~Kisilev, ``Object-to-object color transfer: Optimal flows
  and smsp transformations,'' in \emph{Computer Vision and Pattern Recognition
  (CVPR), 2010 IEEE Conference on}, June 2010, pp. 287--294.

\bibitem{Xiao2009}
X.~Xiao and L.~Ma, ``Gradient-preserving color transfer,'' \emph{Computer
  Graphics Forum}, vol.~28, no.~7, pp. 1879--1886, 2009.

\bibitem{Rabin2011}
J.~Rabin, J.~Delon, and Y.~Gousseau, ``Removing artefacts from color and
  contrast modifications,'' \emph{Image Processing, IEEE Transactions on},
  vol.~20, no.~11, pp. 3073--3085, Nov 2011.

\bibitem{Frigo15}
O.~Frigo, N.~Sabater, V.~Demoulin, and P.~Hellier,
  ``\BIBforeignlanguage{English}{Optimal transportation for example-guided
  color transfer},'' in \emph{\BIBforeignlanguage{English}{Computer Vision --
  ACCV 2014}}, ser. Lecture Notes in Computer Science, D.~Cremers, I.~Reid,
  H.~Saito, and M.-H. Yang, Eds.\hskip 1em plus 0.5em minus 0.4em\relax
  Springer International Publishing, 2015, vol. 9005, pp. 655--670.

\bibitem{Jeong2007}
K.~Jeong and C.~Jaynes, ``Object matching in disjoint cameras using a color
  transfer approach,'' \emph{Machine Vision and Applications}, vol.~19, no.~5,
  pp. 443--455, 2007.

\bibitem{Xiang2009682}
Y.~Xiang, B.~Zou, and H.~Li, ``Selective color transfer with multi-source
  images,'' \emph{Pattern Recognition Letters}, vol.~30, no.~7, pp. 682 -- 689,
  2009.

\bibitem{Comaniciu2002}
D.~Comaniciu and P.~Meer, ``Mean shift: a robust approach toward feature space
  analysis,'' \emph{Pattern Analysis and Machine Intelligence, IEEE
  Transactions on}, vol.~24, no.~5, pp. 603--619, May 2002.

\bibitem{XuICIP2005}
S.~Xu, Y.~Zhang, S.~Zhang, and X.~Ye, ``Uniform color transfer,'' in \emph{IEEE
  International Conference on Image Processing 2005}, vol.~3, Sept 2005, pp.
  III--940--3.

\bibitem{TaiCVPR2005}
Y.-W. Tai, J.~Jia, and C.-K. Tang, ``Local color transfer via probabilistic
  segmentation by expectation-maximization,'' in \emph{2005 IEEE Computer
  Society Conference on Computer Vision and Pattern Recognition (CVPR'05)},
  vol.~1, June 2005, pp. 747--754 vol. 1.

\bibitem{Oliveira2015}
M.~Oliveira, A.~Sappa, and V.~Santos, ``A probabilistic approach for color
  correction in image mosaicking applications,'' \emph{Image Processing, IEEE
  Transactions on}, vol.~24, no.~2, pp. 508--523, Feb 2015.

\bibitem{HaCohen2011}
Y.~HaCohen, E.~Shechtman, D.~B. Goldman, and D.~Lischinski, ``Non-rigid dense
  correspondence with applications for image enhancement,'' \emph{ACM Trans.
  Graph.}, vol.~30, no.~4, pp. 70:1--70:10, Jul. 2011.

\bibitem{Jian2011}
B.~Jian and B.~Vemuri, ``Robust point set registration using gaussian mixture
  models,'' \emph{Pattern Analysis and Machine Intelligence, IEEE Transactions
  on}, vol.~33, no.~8, pp. 1633--1645, Aug 2011.

\bibitem{ArellanoPR2015}
C.~Arellano and R.~Dahyot, ``Robust ellipse detection with gaussian mixture
  models,'' \emph{Pattern Recognition}, 2016.

\bibitem{ITBook2009}
F.~Escolano, P.~Suau, and B.~Bonev, \emph{Information theory in Computer Vision
  and Pattern Recognition}.\hskip 1em plus 0.5em minus 0.4em\relax Springer,
  2009.

\bibitem{ScottTechnometrics2001}
D.~W. Scott, ``\BIBforeignlanguage{English}{Parametric statistical modeling by
  minimum integrated square error},''
  \emph{\BIBforeignlanguage{English}{Technometrics}}, vol.~43, no.~3, pp. pp.
  274--285, 2001.

\bibitem{JianICCV2005}
B.~Jian and B.~C. Vemuri, ``A robust algorithm for point set registration using
  mixture of gaussians,'' in \emph{International Conference on Computer Vision
  (2005)}, 2005.

\bibitem{Eusipco2012Arellano1}
C.~Arellano and R.~Dahyot, ``Shape model fitting algorithm without point
  correspondence,'' in \emph{20th European Signal Processing Conference
  (Eusipco)}, Bucharest, Romania, August, 27-31 2012, pp. 934--938.

\bibitem{CVMP2013Arellano}
------, ``Robust bayesian fitting of 3d morphable model,'' in \emph{Proceedings
  of the 10th European Conference on Visual Media Production}, ser. CVMP
  '13.\hskip 1em plus 0.5em minus 0.4em\relax New York, NY, USA: ACM, 2013, pp.
  9:1--9:10.

\bibitem{BelongiePAMI2002}
S.~Belongie, J.~Malik, and J.~Puzicha, ``Shape matching and object recognition
  using shape contexts,'' \emph{Pattern Analysis and Machine Intelligence, IEEE
  Transactions on}, vol.~24, no.~4, pp. 509--522, Apr 2002.

\bibitem{LoweIJCV2004}
D.~Lowe, ``Distinctive image features from scale-invariant keypoints,''
  \emph{International Journal on Computer Vision}, vol.~60, no.~2, p. 91–110,
  2004.

\bibitem{MaCVPR2013}
J.~Ma, J.~Zhao, J.~Tian, Z.~Tu, and A.~L. Yuille, ``Robust estimation of
  nonrigid transformation for point set registration,'' in \emph{IEEE
  Conference on Computer Vision and Pattern Recognition}, Portland, OR, USA
  USA, June 23-28 2013.

\bibitem{PitieRigidColourTransfer2007}
F.~Piti\'{e} and A.~Kokaram, ``The linear monge-kantorovitch linear (mkl)
  colour mapping for example-based colour transfer,'' in \emph{4th European
  Conference on Visual Media Production}, 2007.

\bibitem{Wu2007}
X.~Wu, V.~Kumar, J.~Ross~Quinlan, J.~Ghosh, Q.~Yang, H.~Motoda, G.~J.
  McLachlan, A.~Ng, B.~Liu, P.~S. Yu, Z.-H. Zhou, M.~Steinbach, D.~J. Hand, and
  D.~Steinberg, ``Top 10 algorithms in data mining,'' \emph{Knowl. Inf. Syst.},
  vol.~14, no.~1, pp. 1--37, Dec. 2007.

\bibitem{Carreira-PAMI07}
M.~Carreira-Perpinan, ``Gaussian mean-shift is an em algorithm,'' \emph{IEEE
  Transactions on Pattern Analysis and Machine Intelligence}, vol.~29, no.~5,
  pp. 767 -- 776, May 2007.

\bibitem{Lissner2013}
I.~Lissner, J.~Preiss, P.~Urban, M.~S. Lichtenauer, and P.~Zolliker,
  ``Image-difference prediction: From grayscale to color,'' \emph{IEEE
  Transactions on Image Processing}, vol.~22, no.~2, pp. 435--446, Feb 2013.

\bibitem{Hristova2015}
H.~Hristova, O.~Le~Meur, R.~Cozot, and K.~Bouatouch, ``Style-aware robust color
  transfer,'' in \emph{Proceedings of the Workshop on Computational
  Aesthetics}, ser. CAE '15.\hskip 1em plus 0.5em minus 0.4em\relax
  Aire-la-Ville, Switzerland, Switzerland: Eurographics Association, 2015, pp.
  67--77.

\bibitem{morovivc2008}
J.~Morovi{\v{c}}, ``Color gamut mapping,'' 2008.

\bibitem{GIMP}
\BIBentryALTinterwordspacing
 [Online]. Available: \url{https://www.gimp.org/}
\BIBentrySTDinterwordspacing

\bibitem{Matlabrgb2ind}
\BIBentryALTinterwordspacing
 [Online]. Available:
  \url{https://uk.mathworks.com/help/matlab/ref/rgb2ind.html}
\BIBentrySTDinterwordspacing

\bibitem{Bonneel2013}
N.~Bonneel, K.~Sunkavalli, S.~Paris, and H.~Pfister, ``Example-based video
  color grading,'' \emph{ACM Trans. Graph.}, vol.~32, no.~4, pp. 39:1--39:12,
  Jul. 2013.

\bibitem{ZhichengACMTOG2016}
Z.~Yan, H.~Zhang, B.~Wang, S.~Paris, and Y.~Yu, ``Automatic photo adjustment
  using deep neural networks,'' \emph{ACM Trans. Graph.}, vol.~35, no.~2, pp.
  11:1--11:15, Feb. 2016.

\end{thebibliography}

\end{document}